\documentclass[10pt,a4paper]{extarticle}

\usepackage{amsmath,amsfonts,bm}

\def\eqref#1{equation~\ref{#1}}
\def\1{\bm{1}}

\DeclareMathAlphabet{\mathsfit}{\encodingdefault}{\sfdefault}{m}{sl}
\SetMathAlphabet{\mathsfit}{bold}{\encodingdefault}{\sfdefault}{bx}{n}

\newcommand{\xsimfull}{Pelican-Sim 1.0}
\newcommand{\papertitle}{\xsimfull{}: A General World Model Simulator \\ for Embodied Intelligence}
\usepackage{graphicx}

\usepackage{float}
\usepackage{placeins}
\usepackage{booktabs}
\usepackage{multirow}
\usepackage{amssymb}
\usepackage[authoryear,round]{natbib}
\usepackage{url}
\usepackage{arpafvg}

\title{\papertitle}
\newcommand*{\projectpageurl}{https://zoushilong1024.github.io/Pelican-Sim1.0/}
\author{%
    Beijing Innovation Center of Humanoid Robotics (X-Humanoid) \\
    \textbf{WFM System Group} \\
    {\fontencoding{T1}\selectfont\{jack.zou,vito.dai,jian.tang,jason.ju\}@x-humanoid.com} \\
    \href{https://github.com/Open-X-Humanoid}{\textcolor{titlecolor}{\texttt{GitHub: Open-X-Humanoid/Pelican-Sim1.0}}} \\
    \ifx\projectpageurl\empty
        \textcolor{titlecolor}{\texttt{Project page: Coming soon}}%
    \else
        \href{\projectpageurl}{\textcolor{titlecolor}{\texttt{Project page: }\expandafter\nolinkurl\expandafter{Pelican-Sim 1.0.github.io}}}%
    \fi
}
\date{September 11, 2026}

\begin{document}
\maketitle
\begin{figure}[H]
    \centering
    \includegraphics[width=\textwidth]{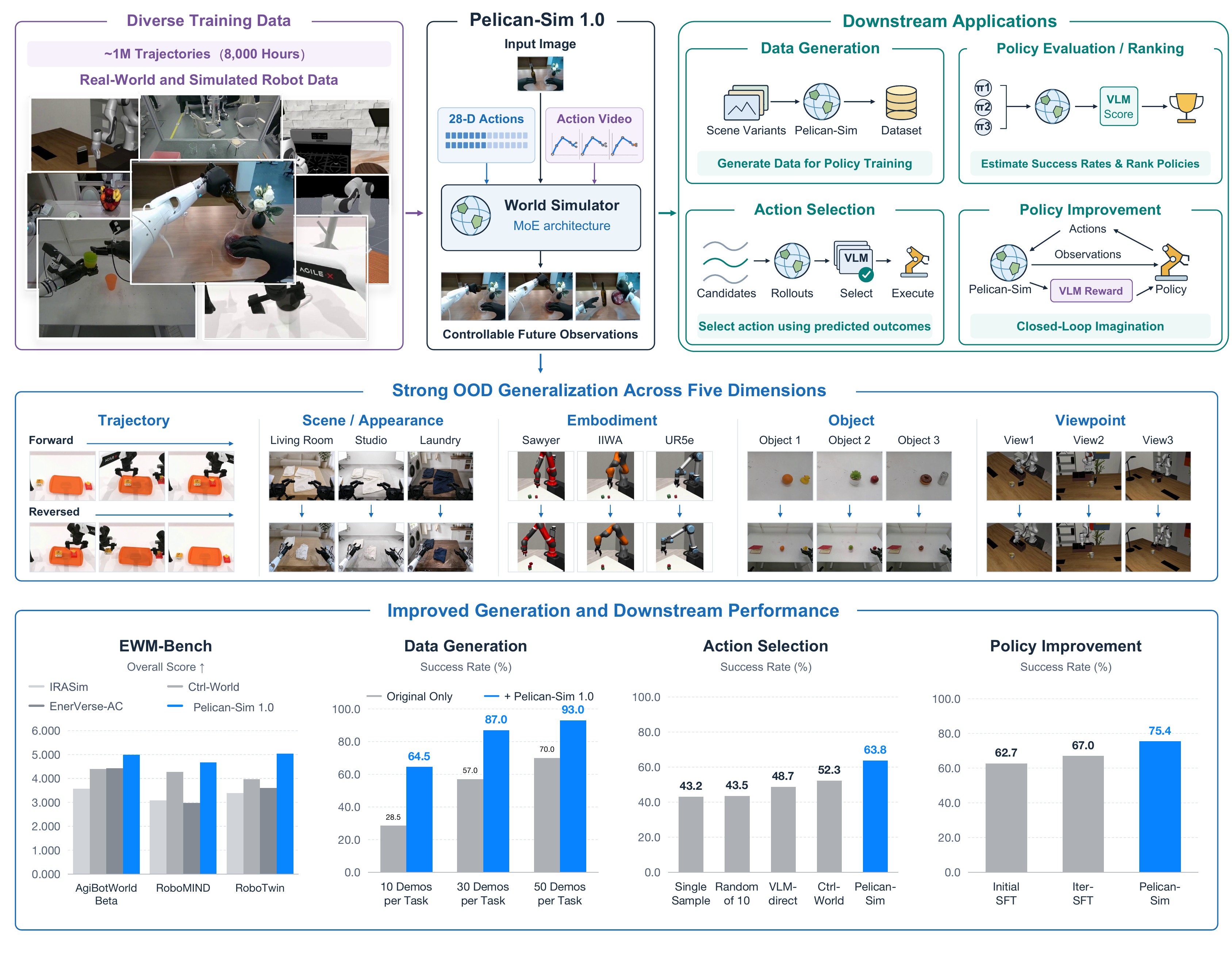}
    \setlength{\abovecaptionskip}{-2pt}
    \caption{
        \textbf{Overview of \xsimfull{}.}
        \textbf{Top:} Trained on approximately one million real-world and simulated trajectories, \xsimfull{} predicts future observations from an initial RGB image, unified 28-dimensional action values, and camera-aligned action videos using a Video DiT with sparse MoE layers. Four downstream applications are illustrated: data generation for policy training, policy evaluation and ranking, action selection, and policy improvement. A fine-tuned VLM evaluator supplies success estimates, candidate scores, and rewards for the latter three applications, respectively; policy improvement uses closed-loop imagined interaction.
        \textbf{Middle:} Qualitative examples illustrate generalization across five dimensions: reversed trajectories, edited scenes and object variants, held-out robot embodiments, and novel viewpoints.
        \textbf{Bottom:} Adapted EWMBench overall scores summarize generation performance. Separate RoboTwin studies evaluate data generation through demonstration augmentation (Data Generation), action selection, and policy improvement against their respective baselines.
    }
    \label{fig:teaser}
\end{figure}
% Figure labels use "Policy Evaluation / Ranking"; prose uses "and ranking".
% The bottom "Data Engine" panel reports the data-generation application.

\section*{Abstract}

In this technical report, we propose \textbf{\xsimfull{}}, a general world model simulator for embodied intelligence that predicts future observations from visual context and robot actions to support downstream learning and decision making.
The model incorporates four key design features: \textbf{(1) Unified action representation}: a 28-dimensional action value space covering most mainstream embodiments, keeping one model valid across heterogeneous devices. \textbf{(2) Action-visual injection}: URDF- and camera-rendered action videos bridge actions and pixels, giving markedly better controllability across embodiments, scenes, and tasks (PSNR $+0.904$ over alternative fusion baselines). \textbf{(3) Sparse mixture-of-experts (MoE)}: sparse MoE layers add capacity for heterogeneous dynamics and absorb the action modality while reducing inter-modality conflict (FVD $-6.530$ vs. the dense backbone). \textbf{(4) Efficient rollout generation}: causal adaptation and few-step distillation yield a four-step autoregressive simulator, achieving a $5.67\times$ speedup over the 35-step model.
\\[0.3em]
Benefiting from these designs, we train on approximately one million real-world and simulated trajectories and obtain large gains in action controllability and video quality: PSNR improves over the strongest evaluated baselines by 4.636 on AgiBotWorld Beta, 2.080 on RoboMIND, and 10.343 on RoboTwin, with the adapted EWMBench DYN score up 0.426 on RoboTwin. Relying on this, four downstream applications on RoboTwin succeed: 500 generated trajectories added to 50 demonstrations per task raise policy success from 70\% to 93\%; policy evaluation reaches a Pearson correlation of 0.994 across five checkpoints; and relative success gains reach 47.7\% for action selection and 20.3\% for policy improvement. Qualitative generalization across trajectory, scene, object, embodiment, and viewpoint shifts highlights its potential as a general-purpose world model simulator.

\section{Introduction}

Action-conditioned world models predict future observations from visual context and candidate robot actions, providing learned simulators for evaluating behavior before execution. Such models can support data generation, policy evaluation, and policy improvement while reducing reliance on additional environment interaction~\citep{yang2024unisim,escontrela2023viper}. Extending this capability across heterogeneous robots requires more than visually realistic prediction: the model must interpret numerical commands, account for embodiment geometry, and predict their effects in diverse scenes. A central challenge is therefore to represent robot motion in a form that preserves precise configuration information while providing transferable geometric guidance.

Existing methods use numerical, visual, or hybrid action interfaces (Table~\ref{tab:action_conditioning_comparison}). IRASim~\citep{zhu2024irasim} and Ctrl-World~\citep{guo2026ctrlworld} condition video generation on low-dimensional actions. Numerical configurations preserve precise joint and gripper states but lack explicit image-space motion guidance~\citep{wang2025vap}; even with a unified layout, the model must learn their visual effects for each robot and camera. Visual conditions make motion explicit in image space but emphasize different information: end-effector maps provide local pose guidance~\citep{jiang2025enerverseac}, embodiment masks encode projected occupancy~\citep{chen2026bridgev2w,gu2026geniworld}, and skeletons expose articulated structure~\citep{wang2025vap,wu2026oscar}. Figure~\ref{fig:visual_action_types} illustrates these geometric cues. End-effector maps leave whole-arm articulation underconstrained, while binary robot masks emphasize projected occupancy rather than explicit joint connectivity. Skeletons expose articulated structure, but, like masks, do not generally determine the underlying numerical configuration uniquely from image-space projections. Hybrid methods such as EnerVerse-AC~\citep{jiang2025enerverseac} and ViPSim~\citep{chen2026vipsim} motivate combining these complementary sources. 
% Our focus is to retain numerical configurations alongside explicit whole-arm guidance within a synchronized interface for cross-embodiment video prediction.
These complementary properties suggest that cross-embodiment action conditioning should jointly preserve precise numerical configurations and explicit image-space motion structure.

To this end, we propose \textbf{\xsimfull{}}, an action-conditioned world model simulator that integrates synchronized numerical and visual action representations. A unified 28-dimensional action value specifies arm-joint and parallel-gripper or dexterous-hand states for single-arm and bimanual robots. The corresponding action video renders whole-arm joint locations and link connectivity using the robot URDF and camera calibration, providing geometric guidance grounded in nominal forward kinematics and camera projection. 
% Separate pathways inject numerical and visual features into alternating blocks of a Diffusion Transformer (DiT). 
The action-value and action-video pathways inject numerical and visual features, respectively, into alternating blocks of the same Video Diffusion Transformer (DiT).
% This design preserves precise configurations while making their projected motion explicit, providing a common conditioning interface for joint training on heterogeneous robot data and action-conditioned generation across embodiments.
By jointly preserving precise configurations and their projected motion structure, this design provides a common conditioning interface for training on heterogeneous robot data and enables action-conditioned video generation across embodiments.

% To accommodate diverse robot embodiments and environments within a unified model, 
The unified action interface makes heterogeneous trajectories compatible as model inputs, but their dynamics still vary across embodiments, tasks, and scenes. We equip the Video DiT with sparse mixture-of-experts (MoE) layers to expand its capacity for modeling heterogeneous robot dynamics~\citep{lepikhin2020gshard,riquelme2021vmoe,fei2024ditmoe,ma2026lingbotvideo}. Routed experts provide transformations selected according to token features, while shared experts provide a common processing pathway across the heterogeneous inputs. We train the resulting model on a curated corpus of approximately one million trajectories spanning real-world and simulated manipulation. Inference acceleration further supports efficient rollout generation without changing the action interface. Together, these components address action representation, modeling capacity, and the computational demands of repeated simulator use. Pelican-Sim 1.0 will serve as a key simulation module in future versions of the Pelican-Unify model family~\citep{zhang2026pelican}.

We conduct a systematic evaluation of \xsimfull{} that progresses from video prediction and design analysis, to data generation, policy evaluation and ranking, action selection, policy improvement, and finally OOD generalization.
Experiments on the held-out test splits of AgiBotWorld Beta~\citep{agibot2025world}, RoboMIND~\citep{wu2024robomind}, and RoboTwin~\citep{mu2024robotwin} show that \xsimfull{} achieves the best results on all five video-quality metrics and the highest adapted EWMBench overall score on each dataset among compared methods with available results. On RoboTwin, for example, it reduces FVD from $26.60$ to $4.98$ relative to Ctrl-World, the strongest baseline on this metric. Ablation studies support the contributions of mixed-domain training, complementary action conditioning, and the MoE architecture. Beyond video prediction, we evaluate data generation through policy training with augmented demonstrations and combine the simulator with a fine-tuned vision language model (VLM) evaluator for policy evaluation and ranking, action selection, and policy improvement. The evaluator supplies task-conditioned scores from which these downstream procedures obtain success estimates, selection utilities, and rewards. Qualitative studies further examine generalization across trajectory, scene appearance, embodiment, object, and viewpoint shifts. Together, these experiments evaluate \xsimfull{} as a general-purpose simulator supporting the full pipeline from future prediction to downstream policy learning and decision making.

% Required packages: booktabs, graphicx, amssymb, multirow

\begin{table*}[t]
\centering

\caption{Comparison of action-conditioning interfaces, the use of token-choice routing in sparse MoE layers, and reported evidence of out-of-distribution (OOD) generalization across robotic world models. OOD entries summarize method-specific evaluations rather than results under a shared protocol.}

\label{tab:action_conditioning_comparison}

\setlength{\tabcolsep}{5pt}
\renewcommand{\arraystretch}{1.12}

\resizebox{\textwidth}{!}{%
\begin{tabular}{lccccccccc}
\toprule

\multirow{2}{*}{\textbf{Method}}
& \multirow{2}{*}{\textbf{Action Value}}
& \multirow{2}{*}{\textbf{Visual Action Type}}
& \multirow{2}{*}{\shortstack{\textbf{Cross-Embod.}\\\textbf{Interface}}}
& \multirow{2}{*}{\shortstack{\textbf{Token-choice}\\\textbf{MoE}}}
& \multicolumn{5}{c}{\textbf{Reported OOD Evidence}} \\
\cmidrule(lr){6-10}
& & & &
& \textbf{Traj.}
& \textbf{Object}
& \textbf{Embod.}
& \textbf{View.}
& \textbf{Scene/App.} \\

\midrule

% Evidence audit (2026-09-07): arXiv:2406.14540, Sec. 4.4/Fig. 9 explicitly
% describes keyboard/VR trajectories as deviating from the training distribution.
% Sec. 7 uses a unified padded 5-D action interface across RoboNet platforms.
IRASim~\citep{zhu2024irasim}
& $\checkmark$
& None
& $\checkmark$
& $\times$
& $\checkmark$
& --
& --
& --
& -- \\

% arXiv:2510.10125, Fig. 4; Sec. 5.3/Fig. 6; Sec. 5.4/Appendix C.
% Direct zero-shot rollouts in a new scene with novel camera placements;
% action variation and novel-object policy improvement provide indirect evidence.
Ctrl-World~\citep{guo2026ctrlworld}
& $\checkmark$
& None
& $\times$
& $\times$
& $\checkmark$
& --
& --
& $\checkmark$
& -- \\

% arXiv:2505.09723, Sec. 3.3: spatially augmented, reversed trajectories
% are used for data synthesis. Multi-view conditioning alone is not viewpoint OOD.
EnerVerse-AC~\citep{jiang2025enerverseac}
& $\checkmark$
& EEF map
& $\times$
& $\times$
& $\checkmark$
& --
& --
& --
& -- \\

% VAP, ICCV 2025, Secs. 3.3 and 4.2: CogVideoX with ControlNet/LoRA;
% one visual interface across RT-1, DROID, and human hands.
% Sec. 3.2/Fig. 3: HOI hand skeletons are estimated from video; robotic
% gripper skeletons are rendered from state logs with optional visual correction.
% Sec. 4 implementation/Table 2: one DROID lab held out for novel scenes.
% Novel-skill evaluation/Figs. 5 and 7 provide only indirect trajectory-OOD
% evidence, not a dedicated trajectory-distribution split. Object consistency
% is not an object-OOD evaluation; multi-agent joint training does not establish
% held-out-embodiment transfer, nor do varied cameras establish viewpoint OOD.
Visual Action Prompts~\citep{wang2025vap}
& $\times$
& Hand/gripper skeleton
& $\checkmark$
& $\times$
& $\checkmark$
& $\triangle$ 
& --
& $\triangle$ 
& $\triangle$  \\

% arXiv:2606.04463, Appendix A.9/Figs. 9--10: held-out Ego4D scenes
% are human-interaction probes, hence indirect evidence for robotic scene OOD.
OSCAR~\citep{wu2026oscar}
& $\times$
& Arm skeleton
& $\checkmark$
& $\times$
& --
& --
& $\triangle$
& --
& -- \\

% arXiv:2602.03793, Sec. 4.1/Table 1: separate unseen-view and unseen-scene
% splits. Training on two embodiments does not establish embodiment OOD.
BridgeV2W~\citep{chen2026bridgev2w}
& $\times$
& Robot mask
& $\checkmark$
& $\times$
& --
& --
& $\triangle$
& $\checkmark$
& $\checkmark$ \\

% arXiv:2608.06332, Sec. IV-A/Table I: Clean-to-Random jointly varies
% object instances, placements, layouts, and scene appearance without adaptation.
% Secs. IV-A/B apply the visual interface to RoboTwin and real Xtrainer robots;
% this is interface reuse, not evidence of transfer to an unseen embodiment.
GeniWorld~\citep{gu2026geniworld}
& $\times$
& Robot mask
& $\checkmark$
& $\times$
& --
& $\checkmark$
& --
& --
& $\checkmark$ \\

% arXiv:2606.28804, Fig. 1: AgiBot-trained model transfers to DROID in OOD
% environments. Figs. 6--7: sparse/action-swap probes; Sec. V explicitly
% reports incomplete object interaction and geometry errors under novel views.
ViPSim~\citep{chen2026vipsim}
& $\checkmark$
& EEF map \& Robot mask
& $\checkmark$
& $\times$
& $\triangle$
& $\triangle$
& $\checkmark$
& $\checkmark$
& $\triangle$ \\

% arXiv:2607.19343, Sec. 5.1/Fig. 5: unseen BEHAVIOR embodiment;
% Appendix D explicitly excludes CLVR/RAD scenes from DROID training.
Masked Visual Actions~\citep{alzayer2026masked}
& $\times$
& Robot mask
& $\checkmark$
& $\times$
& --
& --
& $\checkmark$
& --
& -- \\

\midrule

\textbf{\xsimfull{} (Ours)}
& $\checkmark$
& Whole-arm skeleton (URDF)
& $\checkmark$
& $\checkmark$
& $\checkmark$
& $\checkmark$
& $\checkmark$
& $\checkmark$
& $\checkmark$ \\

\bottomrule
\end{tabular}%
}

\vspace{2pt}

\begin{minipage}{\textwidth}
\footnotesize
Interface/architecture: $\checkmark$ supported; $\triangle$ limited support; $\times$ not used.
OOD: $\checkmark$ directly demonstrated, qualitatively or quantitatively; $\triangle$ limited or indirect evidence; -- not reported, rather than unsupported.
Evidence may involve joint shifts and does not imply comparable robustness.
``Cross-Embod.'' denotes interface reuse, not unseen-robot transfer; ``Scene/App.'' denotes scene or appearance changes.
``Token-choice MoE'' denotes sparse feed-forward expert selection at the token level; this criterion does not require shared experts.
OOD is assessed under each method's reported training and evaluation setup.
\end{minipage}

\vspace{5pt}

\includegraphics[width=\textwidth]{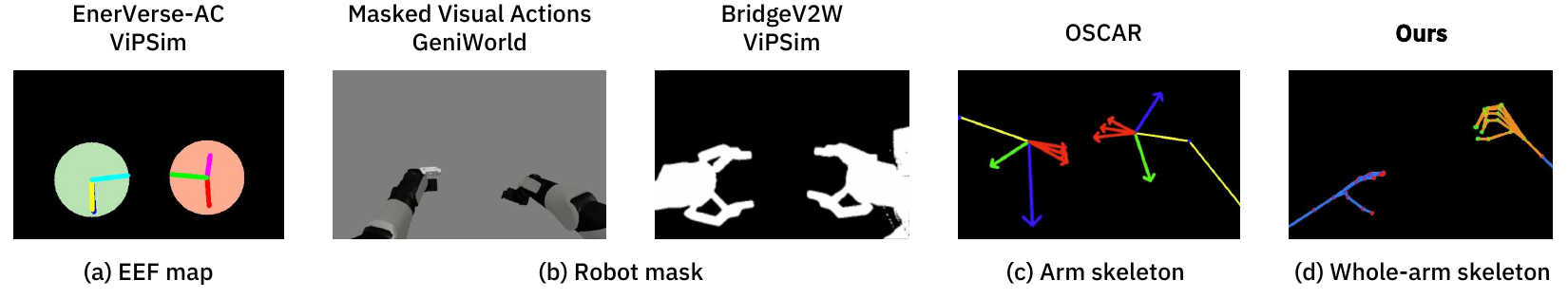}

\vspace{3pt}

\refstepcounter{figure}
\label{fig:visual_action_types}
\begin{minipage}{\textwidth}
\small
\textbf{Figure~\thefigure: Geometric cues in visual action representations.}
\textbf{(a)} End-effector maps provide local pose guidance. \textbf{(b)} Robot masks represent projected occupancy. \textbf{(c)} Arm skeletons depict the kinematic chain and gripper-state cues. \textbf{(d)} Our camera-aligned, URDF-rendered skeleton videos provide whole-arm motion guidance with gripper or hand-state cues. Method names indicate representative approaches.
\end{minipage}

\end{table*}

In summary, our contributions are as follows:
\begin{itemize}
    \item \textbf{Unified cross-embodiment action conditioning.} We introduce \xsimfull{}, a general action-conditioned world model simulator that pairs a unified 28-dimensional numerical action representation with camera-aligned, URDF-rendered whole-arm skeleton videos. Interleaved action-value and action-video pathways integrate precise configuration information and explicit image-space motion geometry within a single Video DiT, enabling joint learning across heterogeneous embodiments. This complementary conditioning improves video prediction over either modality alone; the interleaved architecture improves PSNR by $0.904$ dB over late fusion on AgiBotWorld Beta.

    \item \textbf{Heterogeneous dynamics modeling and efficient rollouts.} We equip the Video DiT with sparse MoE layers that combine shared processing with token-dependent expert selection to increase capacity for heterogeneous robot dynamics. Under the same mixed-domain training setup, this design reduces FVD by $6.530$ relative to the dense backbone on AgiBotWorld Beta. We further combine causal adaptation with few-step distillation to obtain a four-step autoregressive simulator, achieving a $5.67\times$ speedup over the 35-step model in the reported 21-frame benchmark and supporting repeated queries for downstream decision making.

    \item \textbf{Comprehensive evaluation and downstream utility.} Trained on approximately one million real-world and simulated trajectories, \xsimfull{} improves PSNR over the strongest evaluated baselines by $4.636$, $2.080$, and $10.343$ dB on AgiBotWorld Beta, RoboMIND, and RoboTwin, respectively. We demonstrate four downstream applications on RoboTwin: augmenting 50 demonstrations per task with 500 generated trajectories raises policy success from $70\%$ to $93\%$; VLM-assisted policy evaluation achieves a Pearson correlation of $0.994$ across five checkpoints with 1,000 task-specific adaptation rollouts; and action selection and policy improvement yield relative success gains of $47.7\%$ and $20.3\%$ over single-sample execution and supervised initialization, respectively. Qualitative studies across trajectory, scene, object, embodiment, and viewpoint shifts further demonstrate the model's generalization potential.

    \item \textbf{Open-source resources.} We will release \xsimfull{} model checkpoints and inference code to support reproducible evaluation and further research on general-purpose world model simulators for embodied intelligence.
\end{itemize}
\section{Related Work}
\label{sec:related_work}

\subsection{World Models and Action-Conditioned Video Generation}

World models learn predictive dynamics for planning and policy learning. PlaNet~\citep{hafner2019planet} plans from pixels through compact latent dynamics. Video Diffusion Models~\citep{ho2022video} provide a diffusion-based framework for video synthesis, while interactive world models additionally model the effects of actions. DIAMOND~\citep{alonso2024diamond} studies the importance of visual fidelity for downstream control, and Genie~\citep{bruce2024genie} learns interactive environments and latent actions from unlabelled Internet videos.
In robotics, world models have been explored for planning, predictive rewards, and action-conditioned simulation. UniSim~\citep{yang2024unisim} learns an interactive simulator from heterogeneous data with high- and low-level controls. RoboDreamer~\citep{zhou2024robodreamer} factorizes video generation using compositional language structure to support planning with unseen combinations of objects and actions. VIPER~\citep{escontrela2023viper} instead uses video prediction likelihood as a reinforcement-learning reward. For action-conditioned simulation, IRASim~\citep{zhu2024irasim} generates real-robot rollouts, EnerVerse-AC~\citep{jiang2025enerverseac} supports multi-level action conditioning and multi-view generation, and Ctrl-World~\citep{guo2026ctrlworld} combines multi-view prediction with memory for long-horizon rollouts. 

% In contrast to these approaches, our focus within this line of work is an action interface for predictive simulation across heterogeneous robots.
In contrast to these approaches, our work targets end-to-end action-conditioned predictive simulation across heterogeneous robots, using a synchronized numerical--visual action interface to preserve precise configurations and make whole-arm motion explicit in image space, thereby supporting video prediction, data generation, policy evaluation, and policy improvement within a unified framework. 

\subsection{Cross-Embodiment Learning and Visual Action Interfaces}

Cross-embodiment learning seeks to share policy representations across heterogeneous robots. Open X-Embodiment~\citep{openx2024} aggregates robot data for policy learning, while HPT~\citep{wang2024hpt} maps embodiment-specific inputs into tokens for a shared policy backbone. For predictive simulation, a unified numerical interface retains configuration information but leaves the mapping to visible robot motion implicit.
Visual action interfaces address this limitation by providing explicit geometric guidance in image space.
% Visual action interfaces provide explicit image-space guidance. 
Visual Action Prompts (VAP)~\citep{wang2025vap} combines human and robot data using hand skeletons estimated from videos and gripper skeletons rendered from robot state logs. OSCAR~\citep{wu2026oscar} uses URDF-rendered arm skeletons with gripper-state cues, whereas BridgeV2W~\citep{chen2026bridgev2w} and GeniWorld~\citep{gu2026geniworld} render embodiment masks. Masked Visual Actions~\citep{alzayer2026masked} uses partially revealed robot or object trajectories. 
% These representations expose articulated structure or projected occupancy but do not generally preserve the full numerical configuration.
Although these representations make articulated structure or projected occupancy explicit, they do not generally retain the full underlying numerical configuration.
Hybrid approaches therefore combine numerical actions with visual geometric cues.
EnerVerse-AC~\citep{jiang2025enerverseac} and ViPSim~\citep{chen2026vipsim}, for example, combine numerical actions with visual geometry conditions. 

Motivated by this complementarity, \xsimfull{} pairs unified numerical configurations with synchronized, camera-aligned, URDF-rendered whole-arm skeleton videos through interleaved conditioning pathways. This interface combines precise configuration information with explicit motion guidance in a shared model trained across heterogeneous embodiments and domains. Table~\ref{tab:action_conditioning_comparison} and Figure~\ref{fig:visual_action_types} summarize the compared representations.

\subsection{Diffusion Transformers and Sparse Mixture-of-Experts}

Diffusion Transformers (DiTs) use transformer denoisers for scalable diffusion modeling~\citep{peebles2023dit}. Sparse Mixture-of-Experts (MoE) expands feed-forward capacity through conditional expert selection. GShard~\citep{lepikhin2020gshard} studies sparse expert scaling, and V-MoE~\citep{riquelme2021vmoe} applies conditional expert selection to vision transformers. DiT-MoE~\citep{fei2024ditmoe} introduces sparse diffusion transformers with shared expert routing and expert-level balancing, while LingBot-Video~\citep{ma2026lingbotvideo} scales MoE video pretraining for embodied intelligence.

% Building on these architectures, \xsimfull{} replaces dense feed-forward sublayers with sparse MoE layers in a Video DiT conditioned on numerical and visual robot actions. This combines sparse capacity with a unified action interface for modeling real-world and simulated robot interactions.
Building on these architectures, our focus is on modeling heterogeneous robot dynamics under a unified action interface. \xsimfull{} combines a sparse MoE backbone with synchronized numerical and visual action conditions, bringing shared and input-dependent processing to joint training on diverse real-world and simulated robot trajectories.
\section{Method}
\label{sec:method}

\begin{figure*}[t]
    \centering
    \includegraphics[width=\textwidth]{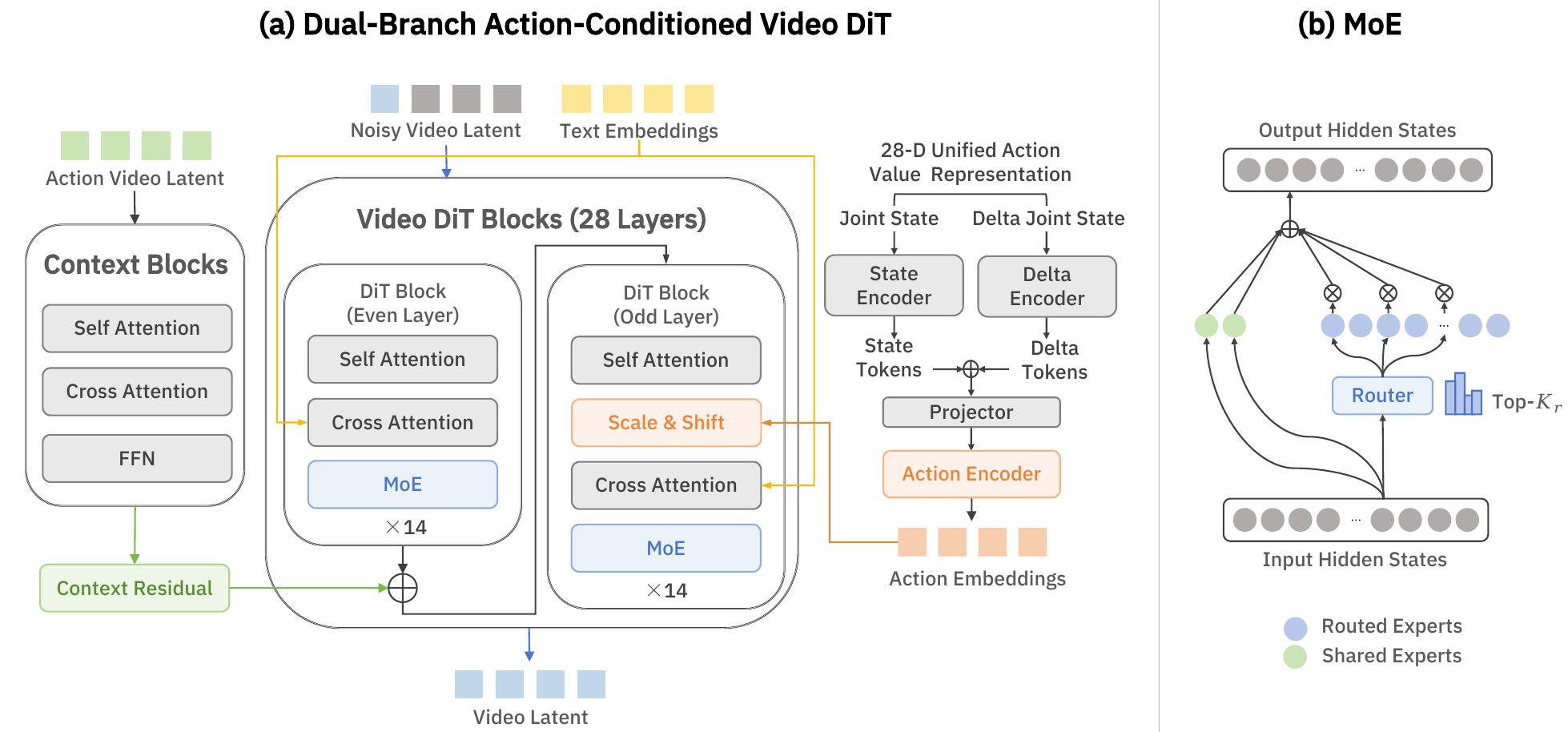}
    \setlength{\abovecaptionskip}{-2pt}
    \caption{
        \textbf{Architecture of \xsimfull{}.}
        \textbf{(a)} A 28-layer Video DiT receives both action conditions: action-value embeddings provide scale-and-shift modulation in odd blocks, while Context Blocks inject action-video residuals at even-block outputs (zero-based indexing).
        \textbf{(b)} Sparse MoE layers combine shared expert outputs with weighted outputs from token-selected routed experts. Expert counts are schematic; adapters, normalization, and gated residual paths are omitted.
    }
    \label{fig:architecture}
\end{figure*}

We present \xsimfull{}, an action-conditioned latent video model that predicts future visual observations from an initial observation and a frame-aligned robot-configuration trajectory. As illustrated in Figure~\ref{fig:architecture}, \xsimfull{} is built on Cosmos-Predict 2.5~\citep{nvidia2025cosmospredict25} and augments its Video DiT with two complementary action-conditioning pathways. A low-dimensional \emph{action value} pathway preserves precise numerical configurations, while an \emph{action video} pathway provides embodiment- and viewpoint-aware geometric guidance. Both action value and action video conditions the same Video DiT: the action-value pathway modulates features within odd-indexed blocks, while the action-video pathway uses auxiliary Context Blocks to supply residuals at even-indexed block outputs. The Video DiT uses sparse MoE layers that combine shared and routed experts.

\subsection{Action-Conditioned Video Generation}
\label{sec:method_formulation}

We consider single-frame conditioning and omit the batch dimension in this subsection. Let $T$ be the number of future frames, $t\in\{0,\ldots,T\}$ the frame index, and $\mathbf{x}_{t}$ the RGB observation at frame $t$. Given the initial observation $\mathbf{x}_{0}$, define
\begin{equation}
    \mathbf{X}^{\mathrm{tar}}
    =(\mathbf{x}_{1},\ldots,\mathbf{x}_{T}),
    \qquad
    \mathbf{A}
    =(\mathbf{a}_{0},\ldots,\mathbf{a}_{T})
    \in\mathbb{R}^{(T+1)\times28}
\end{equation}
as the future video and the frame-aligned robot-configuration trajectory, respectively. Each $\mathbf{a}_{t}\in\mathbb{R}^{28}$ encodes the robot configuration at frame $t$ using the layout in Section~\ref{sec:action_representation}; $\mathbf{a}_{0}$ describes the conditioning frame and $\mathbf{a}_{1:T}$ denotes the subsequence aligned with the target frames. Given an optional task instruction $\mathbf{y}$, the world model with trainable parameters $\theta$ learns the conditional distribution of future observations:
\begin{equation}
    p_{\theta}\!\left(\mathbf{X}^{\mathrm{tar}}
    \mid \mathbf{x}_{0},\mathbf{A},\mathbf{y}\right),
\end{equation}
The robot URDF and camera calibration are known metadata used to construct the action conditions; we omit them from the distribution notation for brevity.

During training, the encoder $\mathcal{E}$ of the pretrained video tokenizer maps the complete clip to a latent sequence. We partition it into the initial-frame conditioning latent $\mathbf{Z}^{\mathrm{cond}}$ and the clean future target latent $\mathbf{Z}^{\mathrm{tar}}_{0}$:
\begin{equation}
    [\mathbf{Z}^{\mathrm{cond}};
     \mathbf{Z}^{\mathrm{tar}}_{0}]
    =\mathcal{E}\!\left(
        [\mathbf{x}_{0};
         \mathbf{X}^{\mathrm{tar}}]
      \right).
\end{equation}
Here, $[\cdot;\cdot]$ concatenates video or latent sequences along time. The subscript $0$ on $\mathbf{Z}^{\mathrm{tar}}_{0}$ denotes the clean endpoint of the flow path, not RGB frame zero. The conditioning latent remains clean, while only the target latent is perturbed. Let $\tau\in[0,1]$ denote the flow timestep, distinct from frame index $t$, and let $\boldsymbol{\epsilon}\sim\mathcal{N}(\mathbf{0},\mathbf{I})$ be standard Gaussian noise with the same shape as the target latent, where $\mathbf{I}$ is the identity covariance after vectorization. We use the linear path
\begin{equation}
    \mathbf{Z}^{\mathrm{tar}}_{\tau}
    =(1-\tau)\mathbf{Z}^{\mathrm{tar}}_{0}
    +\tau\boldsymbol{\epsilon},
    \qquad
    \mathbf{U}^{\mathrm{tar}}_{\tau}
    =\boldsymbol{\epsilon}-\mathbf{Z}^{\mathrm{tar}}_{0},
    \label{eq:flow_path}
\end{equation}
where $\mathbf{U}^{\mathrm{tar}}_{\tau}=\partial\mathbf{Z}^{\mathrm{tar}}_{\tau}/\partial\tau$ is the target velocity. We form the partially noised latent $\widetilde{\mathbf{Z}}_{\tau}=[\mathbf{Z}^{\mathrm{cond}};\mathbf{Z}^{\mathrm{tar}}_{\tau}]$ and use a binary temporal mask $\mathbf{M}$ to identify the clean conditioning positions and noisy target positions. Let $\mathbf{E}^{\mathrm{val}}$ and $\mathbf{Z}^{\mathrm{act}}$ denote the encoded numerical and visual action conditions, constructed in Section~\ref{sec:dual_branch}. The velocity predictor $f_{\theta}$ is trained with the target-only flow-matching objective~\citep{lipman2023flow}:
\begin{equation}
    \mathcal{L}_{\mathrm{flow}}
    =\mathbb{E}_{\mathbf{Z}^{\mathrm{tar}}_{0},
                 \boldsymbol{\epsilon},\tau}
    \left[
    \left\|
    f_{\theta}\!\left(
        \widetilde{\mathbf{Z}}_{\tau},\mathbf{M},\tau
        \mid
        \mathbf{E}^{\mathrm{val}},
        \mathbf{Z}^{\mathrm{act}},
        \mathbf{y}
    \right)_{\mathrm{tar}}
    -\mathbf{U}^{\mathrm{tar}}_{\tau}
    \right\|_{2}^{2}
    \right].
    \label{eq:flow_loss}
\end{equation}
The subscript $\mathrm{tar}$ selects the predictor outputs at future latent positions, and $\|\cdot\|_2$ is the Euclidean norm over vectorized target entries. The expectation is over training examples with their paired conditions, Gaussian noise, and sampled flow timesteps. At inference, we keep $\mathbf{Z}^{\mathrm{cond}}=\mathcal{E}(\mathbf{x}_{0})$ fixed and integrate the learned velocity field from noise at $\tau=1$ to the generated target latent $\widehat{\mathbf{Z}}^{\mathrm{tar}}_{0}$ at $\tau=0$. The tokenizer decoder reconstructs the complete clip from $[\mathbf{Z}^{\mathrm{cond}};\widehat{\mathbf{Z}}^{\mathrm{tar}}_{0}]$; discarding its conditioning frame yields the predicted future video $\widehat{\mathbf{X}}^{\mathrm{tar}}$. Hats on video and latent variables denote generated predictions.
% TODO: Specify the flow-timestep sampling distribution and mask convention.
% TODO: Confirm that initial-frame tokenization is independent of future RGB
% frames and that the training and inference conditioning latents agree.

\subsection{Complementary Action Representations}
\label{sec:action_representation}

\xsimfull{} represents the robot-configuration trajectory in two synchronized forms. The action value preserves precise numerical configuration information in a unified layout, while the action video exposes projected joint locations and link connectivity using the corresponding robot URDF and camera calibration. These are complementary representations of the same trajectory, rather than independently specified controls: the numerical branch retains configuration information that image projection may obscure, and the visual branch makes whole-arm motion explicit in image space. Both representations span the complete $T+1$-frame clip: their first element aligns with the conditioning frame and the remaining elements align with the target frames.

\paragraph{Unified action value.}
To accommodate single-arm and bimanual robots equipped with either parallel grippers or dexterous hands, we pack each frame-aligned robot configuration into a fixed 28-dimensional bilateral layout. The first 14 dimensions describe the left side and the remaining 14 dimensions describe the right side. For side $b\in\{L,R\}$, denoting left or right, we define
\begin{equation}
    \mathbf{a}^{b}_{t}
    =\left[
        \mathbf{q}^{\mathrm{arm},b}_{t};
        g^{b}_{t};
        \mathbf{q}^{\mathrm{hand},b}_{t}
    \right]
    \in\mathbb{R}^{14},
    \label{eq:side_action_value}
\end{equation}
where $\mathbf{q}^{\mathrm{arm},b}_{t}\in\mathbb{R}^{7}$ contains the arm joint angles, $g^{b}_{t}\in\mathbb{R}$ is the parallel-gripper opening, and $\mathbf{q}^{\mathrm{hand},b}_{t}\in\mathbb{R}^{6}$ contains the dexterous-hand joint values. For configuration vectors, $[\cdot;\cdot]$ denotes concatenation along the feature dimension. The complete action value is ordered as
\begin{equation}
    \mathbf{a}_{t}
    =\left[
        \mathbf{a}^{L}_{t};
        \mathbf{a}^{R}_{t}
    \right]
    =\left[
        \underbrace{\mathbf{q}^{\mathrm{arm},L}_{t},
        g^{L}_{t},
        \mathbf{q}^{\mathrm{hand},L}_{t}}_{14\ \mathrm{dimensions}};
        \underbrace{\mathbf{q}^{\mathrm{arm},R}_{t},
        g^{R}_{t},
        \mathbf{q}^{\mathrm{hand},R}_{t}}_{14\ \mathrm{dimensions}}
    \right]
    \in\mathbb{R}^{28}.
    \label{eq:action_value}
\end{equation}
For a 6-DoF arm, the unused seventh arm dimension is set to zero. Missing grippers or dexterous hands are represented by zero-filled slots, and all 14 dimensions of an absent side are set to zero for single-arm embodiments. This canonical ordering preserves the original numerical commands while presenting a fixed action interface across heterogeneous robot embodiments.

\begin{figure*}[t]
    \centering
    \includegraphics[width=\textwidth]{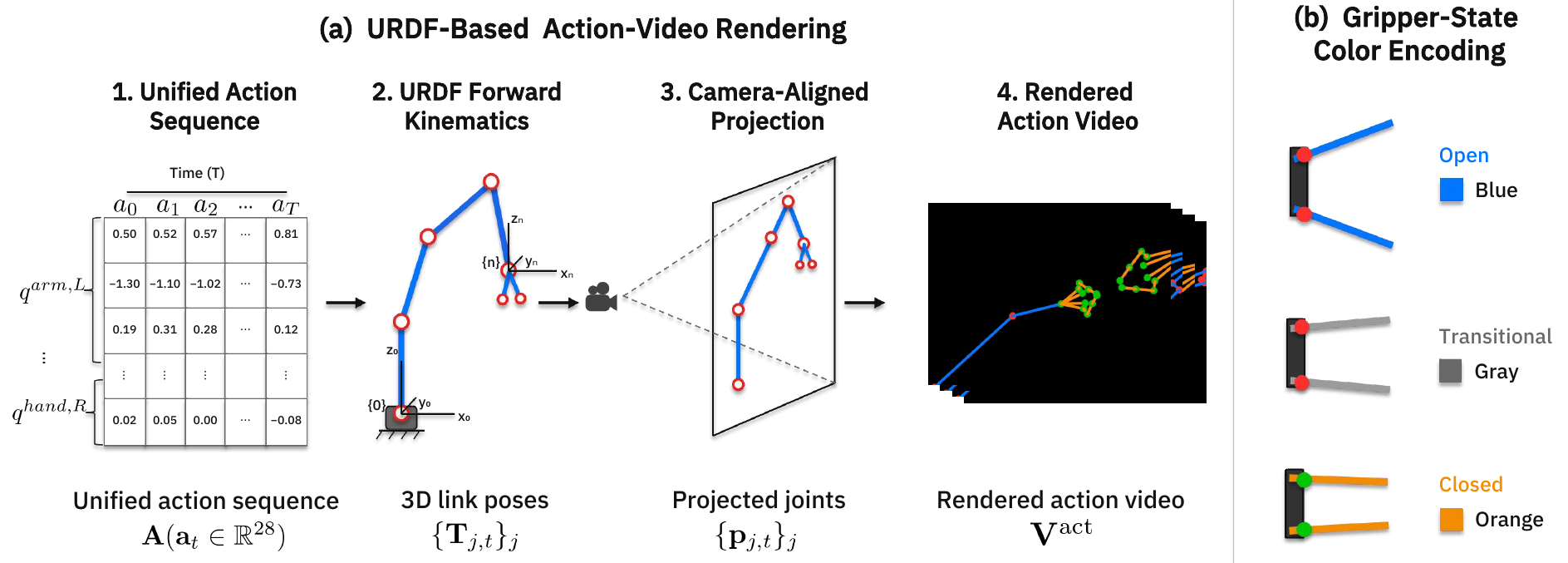}
    \setlength{\abovecaptionskip}{-2pt}
    \caption{
        \textbf{URDF-based action-video rendering.}
        \textbf{(a)} The frame-aligned configuration sequence is mapped to robot-specific URDF joint configurations, converted into 3D link poses through forward kinematics, projected using the paired RGB camera calibration, and rasterized as a synchronized skeleton action video. \textbf{(b)} Gripper links are color-coded to indicate open, transitional, and closed states.
    }
    \label{fig:action_video_rendering}
\end{figure*}

\paragraph{URDF-rendered action video.}
Although the unified action values specify robot configurations precisely, their fixed-dimensional layout does not explicitly describe the resulting image-space motion. Without a visual action condition, the model must learn the mapping from numerical configurations to visible robot motion. We therefore use the robot URDF and camera calibration to render each numerical action sequence as a skeleton video aligned with the RGB view, making this geometric mapping explicit in the conditioning input. As illustrated in Figure~\ref{fig:action_video_rendering}, the unified action at time $t$ is first mapped to the joint configuration required by the corresponding URDF $\mathcal{U}$:
\begin{equation}
    \mathbf{q}^{\mathcal{U}}_{t}
    =\Psi_{\mathcal{U}}(\mathbf{a}_{t}),
    \label{eq:urdf_action_mapping}
\end{equation}
where $\mathbf{q}^{\mathcal{U}}_{t}$ is the configuration in URDF joint order and $\Psi_{\mathcal{U}}$ is the robot-specific conversion from the unified layout. For rendered link index $j$, the forward-kinematics function $\operatorname{FK}_{j}$ returns the homogeneous link-pose transform $\mathbf{T}_{j,t}\in\mathbb{R}^{4\times4}$:
\begin{equation}
    \mathbf{T}_{j,t}
    =\operatorname{FK}_{j}
    (\mathcal{U},\mathbf{q}^{\mathcal{U}}_{t}).
    \label{eq:action_video_fk}
\end{equation}
Let $\mathbf{K}\in\mathbb{R}^{3\times3}$ and $\mathbf{T}_{\mathrm{cam}}\in\mathbb{R}^{4\times4}$ denote the intrinsics and reference-to-camera transform of the paired RGB view, with all link poses expressed in the same reference frame. The link origins are projected onto the image plane as
\begin{equation}
    \mathbf{p}_{j,t}
    =\Pi\!\left(
        \mathbf{K},
        \mathbf{T}_{\mathrm{cam}},
        \mathbf{T}_{j,t}
    \right),
    \label{eq:action_camera_projection}
\end{equation}
where $\Pi$ denotes perspective projection and $\mathbf{p}_{j,t}\in\mathbb{R}^{2}$ is the image coordinate of link origin $j$. Let $\mathcal{K}_{\mathcal{U}}$ be the URDF kinematic edge set connecting these projected nodes. The rasterizer $\mathcal{R}$ produces an action-video frame $\mathbf{v}^{\mathrm{act}}_{t}$ and the full action video $\mathbf{V}^{\mathrm{act}}$:
\begin{equation}
    \mathbf{v}^{\mathrm{act}}_{t}
    =\mathcal{R}\!\left(
        \{\mathbf{p}_{j,t}\}_{j},
        \mathcal{K}_{\mathcal{U}},
        \mathbf{g}_{t}
    \right),
    \qquad
    \mathbf{V}^{\mathrm{act}}
    =(\mathbf{v}^{\mathrm{act}}_{0},\ldots,
      \mathbf{v}^{\mathrm{act}}_{T}),
    \label{eq:action_video_render}
\end{equation}
where $\mathbf{g}_{t}=(g^{L}_{t},g^{R}_{t})$ contains the gripper openings used for state coloring. Arm links and joints form a texture-free skeleton, while gripper links use different colors to distinguish open, transitional, and closed states. This representation provides image-space motion guidance consistent with the nominal robot kinematics and camera projection, while leaving scene appearance to the RGB condition and the generative backbone. This consistency applies to the rendered condition; it does not impose a hard kinematic constraint on the generated RGB observations.
By construction, $\mathbf{v}^{\mathrm{act}}_{t}$ is synchronized with RGB frame $\mathbf{x}_{t}$. The first action-video frame anchors the initial robot configuration, while $\mathbf{v}^{\mathrm{act}}_{1:T}$ provides frame-aligned geometric guidance for $\mathbf{X}^{\mathrm{tar}}$.

\subsection{Dual-Branch Conditioning and Sparse MoE Layers}
\label{sec:dual_branch}

The Video DiT follows the 28-layer latent video architecture of Cosmos-Predict 2.5~\citep{nvidia2025cosmospredict25}. In addition to $\widetilde{\mathbf{Z}}_{\tau}$, $\mathbf{M}$, $\tau$, and optional instruction $\mathbf{y}$, it receives both action-value and action-video conditions through separate pathways. Both conditions act on the same sequence of Video DiT hidden states; the Context Blocks form an auxiliary action-video conditioning branch. Using zero-based block indices $\ell\in\{0,\ldots,27\}$, action values modulate the 14 odd-indexed blocks $\mathcal{I}_{\mathrm{val}}=\{1,3,\ldots,27\}$, while action-video features enter the 14 even-indexed blocks $\mathcal{I}_{\mathrm{vid}}=\{0,2,\ldots,26\}$ through a parallel context branch. Let $\mathbf{H}^{\ell}$ and $\mathbf{H}^{\ell+1}$ denote the input and output hidden features of block $\ell$ in the Video DiT, and let $\mathbf{H}^{\ell}_{\mathrm{moe}}$ denote its features after the MoE layer and gated residual update, before any action-video residual is added.

We now include the batch dimension $B$. Let $T_{\mathrm{lat}}$ be the complete latent sequence length, including the conditioning position, $H_{\mathrm{tok}}\times W_{\mathrm{tok}}$ the spatial token grid after patch embedding, and $D$ the DiT hidden width. Hidden features and aligned action embeddings are represented on the grid $B\times T_{\mathrm{lat}}\times H_{\mathrm{tok}}\times W_{\mathrm{tok}}\times D$, with an equivalent flattened token representation used inside transformer blocks.

\paragraph{Action-value encoding and injection.}
As shown in Figure~\ref{fig:architecture}(a), the action-value pathway represents the trajectory by its initial configuration $\mathbf{s}$ and scaled frame-to-frame changes $\boldsymbol{\delta}_{t}$:
\begin{equation}
    \mathbf{s}=\mathbf{a}_{0},
    \qquad
    \boldsymbol{\delta}_{t}
    =\left(\mathbf{a}_{t}-\mathbf{a}_{t-1}\right)
     \odot\boldsymbol{\eta},
    \quad t=1,\ldots,T,
    \label{eq:action_state_delta}
\end{equation}
where $\boldsymbol{\eta}\in\mathbb{R}^{28}$ is a dimension-wise scaling vector and $\odot$ denotes element-wise multiplication. Let $r_t$ be the tokenizer's temporal compression factor and $T_{\mathrm{a}}=1+\lceil T/r_t\rceil$ the number of action tokens. We group the deltas into $T_{\mathrm{a}}-1$ non-overlapping chunks of length $r_t$, padding the final chunk when necessary. Two multilayer perceptrons (MLPs), $\mathcal{M}_{\mathrm{state}}$ and $\mathcal{M}_{\mathrm{delta}}$, encode the initial state and the delta chunks:
\begin{equation}
    \mathbf{e}^{\mathrm{state}}
    =\mathcal{M}_{\mathrm{state}}(\mathbf{s}),
    \qquad
    \mathbf{e}^{\mathrm{delta}}_{k}
    =\mathcal{M}_{\mathrm{delta}}\!\left(
        \operatorname{Concat}
        (\boldsymbol{\delta}_{kr_t+1:(k+1)r_t})
    \right),
    \quad k=0,\ldots,T_{\mathrm{a}}-2,
    \label{eq:action_chunk_encoding}
\end{equation}
Here, $\mathbf{e}^{\mathrm{state}}$ is the initial-state token, $\mathbf{e}^{\mathrm{delta}}_{k}$ is delta-chunk token $k$, and $\operatorname{Concat}$ concatenates the chunk's $r_t$ delta vectors along the feature dimension. A learned projector and residual action encoder align these tokens to $T_{\mathrm{lat}}$ latent positions and width $D$. Spatial broadcasting then yields $\mathbf{E}^{\mathrm{val}}\in\mathbb{R}^{B\times T_{\mathrm{lat}}\times H_{\mathrm{tok}}\times W_{\mathrm{tok}}\times D}$.
% TODO: Report the delta scaling values, final-chunk padding rule, and the
% temporal alignment/projector implementation, including the value of r_t.

For each odd block $\ell\in\mathcal{I}_{\mathrm{val}}$ of the Video DiT, a block-specific residual adapter transforms the shared action embeddings before producing the scale $\boldsymbol{\gamma}^{\ell}$ and shift $\boldsymbol{\beta}^{\ell}$. These parameters modulate the post-self-attention feature $\overline{\mathbf{H}}^{\ell}$ before text cross-attention:
\begin{equation}
\begin{aligned}
    \mathbf{E}^{\mathrm{inj},\ell}
    &=\mathbf{E}^{\mathrm{val}}
      +\mathcal{I}^{\ell}_{2}\!\left(
        \operatorname{SiLU}\!\left(
            \mathcal{I}^{\ell}_{1}
            (\mathbf{E}^{\mathrm{val}})
        \right)
      \right), \\
    (\boldsymbol{\gamma}^{\ell},\boldsymbol{\beta}^{\ell})
    &=\mathcal{S}^{\ell}
      (\mathbf{E}^{\mathrm{inj},\ell}), \\
    \widehat{\mathbf{H}}^{\ell}
    &=\left(\mathbf{1}+\boldsymbol{\gamma}^{\ell}\right)
      \odot\overline{\mathbf{H}}^{\ell}
      +\boldsymbol{\beta}^{\ell},
    \qquad \ell\in\mathcal{I}_{\mathrm{val}}.
\end{aligned}
    \label{eq:action_film}
\end{equation}
Here, $\mathbf{E}^{\mathrm{inj},\ell}$ is the block-adapted action embedding; $\mathcal{I}^{\ell}_{1}$ and $\mathcal{I}^{\ell}_{2}$ are adapter layers with the SiLU activation between them, and $\mathcal{S}^{\ell}$ is the scale--shift head. The all-ones tensor $\mathbf{1}$ supplies the identity scale, and $\widehat{\mathbf{H}}^{\ell}$ denotes the modulated hidden feature. Scale and shift are aligned with the hidden-feature grid. The modulated feature passes through text cross-attention and the MoE layer, yielding $\mathbf{H}^{\ell+1}=\mathbf{H}_{\mathrm{moe}}^{\ell}$ for odd blocks.

\paragraph{Action-video encoding and injection.}
As shown in Figure~\ref{fig:architecture}(a), let $\overline{\mathbf{V}}^{\mathrm{act}}$ denote the action video normalized to $[-1,1]$. The same video tokenizer $\mathcal{E}$ used for RGB observations produces
\begin{equation}
    \mathbf{Z}^{\mathrm{act}}
    =\mathcal{E}(\overline{\mathbf{V}}^{\mathrm{act}})
    \in\mathbb{R}^{B\times C_z\times T_{\mathrm{lat}}\times H'\times W'},
    \qquad C_z=16,
    \label{eq:action_video_encoding}
\end{equation}
where $C_z$ is the tokenizer's latent-channel dimension and $H'\times W'$ is the latent spatial grid before patch embedding, distinct from the token grid $H_{\mathrm{tok}}\times W_{\mathrm{tok}}$. After patch embedding and spatial alignment, $\mathbf{Z}^{\mathrm{act}}$ is processed by parallel Context Blocks initialized from the corresponding pretrained Video DiT blocks. Context Block $k\in\{0,\ldots,13\}$ produces a residual $\mathbf{R}^{k}$ with the same shape as the Video DiT hidden features. It is added to the output of even block $\ell=2k$ after the MoE layer and its gated residual update:
\begin{equation}
    \mathbf{H}^{\ell+1}
    =\mathbf{H}_{\mathrm{moe}}^{\ell}
     +\mathbf{R}^{\ell/2},
    \qquad \ell\in\mathcal{I}_{\mathrm{vid}},
    \label{eq:action_video_injection}
\end{equation}
Thus, action value and action video are injected into the same Video DiT at alternating blocks: numerical features modulate intermediate states within odd blocks, while visual features contribute spatially and temporally aligned residuals after even blocks. The Context Blocks provide visual conditioning rather than a separate video-prediction backbone.

\paragraph{Sparse MoE layers.}
As illustrated in Figure~\ref{fig:architecture}(b), we replace the dense feed-forward network (FFN) in each block of the Video DiT with a token-choice sparse MoE layer~\citep{fei2024ditmoe,ma2026lingbotvideo}. Let $N_s$ and $N_r$ denote the numbers of shared and routed experts, respectively, and let $K_r\leq N_r$ be the number of routed experts selected per token~\citep{dai2024deepseekmoe}. Each token passes through all shared experts and its top-$K_r$ routed experts. The shared branch provides a common transformation, while the routed branch supplies a token-dependent combination of expert outputs. Expert selection depends on token features rather than predefined embodiment or domain assignments. We use $N_s=1$, $N_r=8$, and $K_r=2$ in all experiments.

Let $n$ index flattened spatiotemporal tokens, and let $\mathbf{h}^{\ell,\mathrm{ca}}_{n}\in\mathbb{R}^{D}$ be token $n$ after text cross-attention in block $\ell$. The router and experts operate on its adaptive layer-normalized (AdaLN) representation $\mathbf{u}^{\ell}_{n}\in\mathbb{R}^{D}$, conditioned on flow timestep $\tau$. Following sigmoid token-choice routing with selection bias~\citep{deepseekai2024deepseekv3,ma2026lingbotvideo}, we define
\begin{equation}
\begin{aligned}
    \mathbf{u}^{\ell}_{n}
    &=\operatorname{AdaLN}^{\ell}_{\mathrm{mlp}}
      (\mathbf{h}^{\ell,\mathrm{ca}}_{n};\tau), \\
    \boldsymbol{\rho}^{\ell}_{n}
    &=\operatorname{sigmoid}
      (\mathbf{W}^{\ell}_{r}\mathbf{u}^{\ell}_{n}), \\
    \mathcal{J}^{\ell}_{n}
    &=\operatorname{TopK}_{K_r}\!\left(
        \boldsymbol{\rho}^{\ell}_{n}+\mathbf{b}^{\ell}_{r}
      \right), \\
    g^{\ell}_{i,n}
    &=\kappa_r
      \frac{\rho^{\ell}_{i,n}}
      {\sum_{j\in\mathcal{J}^{\ell}_{n}}\rho^{\ell}_{j,n}},
    \quad i\in\mathcal{J}^{\ell}_{n}.
\end{aligned}
    \label{eq:router}
\end{equation}
Here, $\mathbf{W}^{\ell}_{r}\in\mathbb{R}^{N_r\times D}$ is the router projection, $\boldsymbol{\rho}^{\ell}_{n}\in\mathbb{R}^{N_r}$ contains its element-wise sigmoid affinities, and $\mathbf{b}^{\ell}_{r}\in\mathbb{R}^{N_r}$ is the selection bias. The operator $\operatorname{TopK}_{K_r}$ returns the index set $\mathcal{J}^{\ell}_{n}$ of the $K_r$ highest-scoring experts. For routed expert index $i$, $\rho^{\ell}_{i,n}$ is the corresponding affinity and $g^{\ell}_{i,n}$ is its mixture weight, normalized over the selected experts and scaled by $\kappa_r$. Once the expert set is selected, mixture weights use the unbiased affinities.
% TODO: Specify the selection-bias initialization/update rule and its relation
% to the auxiliary balancing objective; do not assume an auxiliary-loss-free router.

The MoE layer returns a feed-forward update, while the original gated residual remains outside the module:
\begin{equation}
\begin{aligned}
    \Delta\mathbf{h}^{\ell}_{n}
    &=\sum_{m=1}^{N_s}
        \mathcal{F}^{\mathrm{shared},\ell}_{m}
        (\mathbf{u}^{\ell}_{n})
      +s_r(q)
       \sum_{i\in\mathcal{J}^{\ell}_{n}}
        g^{\ell}_{i,n}
        \mathcal{F}^{\mathrm{routed},\ell}_{i}
        (\mathbf{u}^{\ell}_{n}), \\
    \mathbf{h}^{\ell,\mathrm{moe}}_{n}
    &=\mathbf{h}^{\ell,\mathrm{ca}}_{n}
      +\mathbf{g}^{\ell}_{\mathrm{mlp}}
       \odot\Delta\mathbf{h}^{\ell}_{n}.
\end{aligned}
    \label{eq:moe}
\end{equation}
Here, $\mathcal{F}^{\mathrm{shared},\ell}_{m}$ and $\mathcal{F}^{\mathrm{routed},\ell}_{i}$ are FFNs mapping $\mathbb{R}^{D}$ to $\mathbb{R}^{D}$, with $m$ indexing shared experts and $i$ indexing routed experts. At training iteration $q$, the multiplier $s_r(q)\in[0,1]$ ramps up the routed branch as specified in Section~\ref{sec:experimental_setup}. The vector $\mathbf{g}^{\ell}_{\mathrm{mlp}}\in\mathbb{R}^{D}$ is the original DiT's adaptive residual gate, distinct from expert mixture weights $g^{\ell}_{i,n}$. The update $\Delta\mathbf{h}^{\ell}_{n}$ produces the residual-stream token $\mathbf{h}^{\ell,\mathrm{moe}}_{n}$; collecting and reshaping these tokens gives $\mathbf{H}^{\ell}_{\mathrm{moe}}$ used in both injection pathways.
% TODO: Report expert hidden widths, expert/router initialization, and total
% versus active parameter counts for the Dense and MoE variants.

For load balancing~\citep{fedus2022switch,deepseekai2024deepseekv3}, let $f^{\ell}_{i}$ and $P^{\ell}_{i}$ denote the assignment fraction and mean normalized routing affinity of routed expert $i$ in layer $\ell$. With $L_{\mathrm{M}}=28$ MoE layers and auxiliary-loss weight $\lambda_{\mathrm{MoE}}$, the balancing loss $\mathcal{L}_{\mathrm{MoE}}$ and total training objective $\mathcal{L}$ are
\begin{equation}
    \mathcal{L}_{\mathrm{MoE}}
    =\frac{\lambda_{\mathrm{MoE}}}{L_{\mathrm{M}}}
      \sum_{\ell=0}^{L_{\mathrm{M}}-1}
      N_r\sum_{i=1}^{N_r}f^{\ell}_{i}P^{\ell}_{i},
    \qquad
    \mathcal{L}
    =\mathcal{L}_{\mathrm{flow}}+\mathcal{L}_{\mathrm{MoE}}.
    \label{eq:moe_balance}
\end{equation}
Let $N_{\mathrm{tok}}$ be the number of tokens included in the routing statistics and $\mathbb{I}[\cdot]$ the indicator function, equal to one when its condition holds and zero otherwise. Then $f^{\ell}_{i}=(N_{\mathrm{tok}}K_r)^{-1}\sum_{n=1}^{N_{\mathrm{tok}}}\mathbb{I}[i\in\mathcal{J}^{\ell}_{n}]$ is normalized over all top-$K_r$ assignments, and $P^{\ell}_{i}=N_{\mathrm{tok}}^{-1}\sum_{n=1}^{N_{\mathrm{tok}}}\rho^{\ell}_{i,n}/\sum_{j=1}^{N_r}\rho^{\ell}_{j,n}$ averages affinities normalized over all routed experts. Thus, both $\sum_i f^{\ell}_{i}$ and $\sum_i P^{\ell}_{i}$ equal one.
% TODO: Report lambda_MoE and whether routing statistics are aggregated per
% sequence, per minibatch, or across devices.

\subsection{Causal Few-Step Distillation for Efficient Rollout}
\label{sec:inference_acceleration}

The bidirectional model prioritizes generation quality, but its 35-step sampler is costly for downstream applications that require large numbers of rollouts. We therefore derive a four-step causal student while preserving the task, action-value, and action-video interfaces of \xsimfull{}. Following recent streaming world models~\citep{cheng2026m4world,huang2025selfforcing,zhu2026causalforcing}, we adopt a progressive procedure that separates causal adaptation from few-step distillation.

\paragraph{Progressive causal distillation.}
We partition the target latent video into temporal chunks and initialize the causal model from the bidirectional \xsimfull{} weights. Full temporal self-attention is replaced by block-causal attention, such that each chunk attends only to itself and preceding visual chunks, while the known action conditions for the current chunk remain fully accessible. We first perform teacher-forced flow matching using clean historical chunks, producing a 35-step causal teacher. Its probability-flow ODE trajectories are then used to initialize a four-step causal student~\citep{zhu2026causalforcing}. Finally, we expose the student to its own autoregressive history and apply distribution matching distillation~\citep{yin2024dmd,huang2025selfforcing} against the frozen bidirectional teacher. A small fraction of teacher-forced updates is retained during this self-forcing stage to stabilize training and preserve action controllability.

\paragraph{Accelerated streaming inference.}
During inference, the student generates each latent chunk in four denoising steps and appends it to a rolling context window. Block-causal attention enables key-value caching, avoiding repeated computation over historical tokens. In a 21-frame inference benchmark, the accelerated model requires $2.2$ seconds per clip and achieves a measured $5.67\times$ speedup over the original 35-step model. We use this accelerated four-step causal model for all downstream applications in Sections~\ref{sec:data_engine}--\ref{sec:policy_improvement}.
% TODO: Specify GPU model/count, inference batch size, precision, output
% resolution, timing scope (e.g., encoding, decoding, action rendering,
% serialization, and cache warm-up), and timing statistics for this benchmark.
% Clarify whether the 21 frames include the initial conditioning frame, and
% confirm the baseline configuration used for the reported 5.67x speedup.
% TODO: Specify latent chunk length, rolling-context length, and training
% budgets for the causal adaptation and distillation stages.
\section{Experiments}
\label{sec:experiments}
% Experiment panels use the full column width in both publication formats.
% Tall source sheets use concise captions instead of being globally downscaled.

% We evaluate \xsimfull{} through comparisons with existing world models, component ablations, downstream simulator applications, and generalization under distribution shifts.
% Our experiment follows a progressive evaluation from \textbf{prediction quality}, to \textbf{data utility}, to \textbf{policy evaluation}, and finally to \textbf{action selection}. Specifically, Sections 4.2--4.4 establish predictive performance, Section 4.5 studies the utility of generated rollouts for policy learning, Section 4.6 evaluates their use for policy assessment and ranking, and Section 4.7 examines whether predicted futures can guide candidate action selection.
Our experiments follow a progressive evaluation from \textbf{prediction quality}, to \textbf{data utility}, \textbf{policy evaluation}, \textbf{action selection}, \textbf{policy improvement}, and finally \textbf{OOD generalization}. Specifically, Sections 4.2--4.4 establish predictive performance, Section 4.5 studies the utility of generated rollouts for policy learning, Section 4.6 evaluates their use for policy assessment and ranking, Section 4.7 examines whether predicted futures can guide candidate action selection, Section 4.8 investigates whether closed-loop imagined rollouts can directly improve policies through reinforcement learning, and Section 4.9 assesses robustness under controlled distribution shifts in trajectory, scene appearance, embodiment, object, and viewpoint.

\subsection{Experimental Setup}
\label{sec:experimental_setup}

\paragraph{Training datasets.}
Our training corpus combines real-world demonstrations from AgiBotWorld Beta~\citep{agibot2025world}, RealSource World~\citep{realman2025realsource}, and RoboMIND~\citep{wu2024robomind} with simulated trajectories from LIBERO~\citep{liu2023libero}, ManiSkill2~\citep{gu2023maniskill2}, RoboTwin~\citep{mu2024robotwin}, and RoboCasa~\citep{nasiriany2024robocasa}. These sources cover single-arm and bimanual manipulation with parallel grippers and dexterous hands. All trajectories are converted into unified action values and camera-aligned action videos as described in Section~\ref{sec:action_representation}.

\paragraph{Data quality curation.}
We discard trajectories with incomplete observations, invalid actions, missing or invalid URDF files, missing camera calibration, or inconsistent timestamps. A vision--language model checks the rendered action videos against synchronized RGB observations for whole-arm pose, motion direction, gripper state, temporal alignment, and robot visibility. Low-scoring samples undergo automatic alignment correction, rerendering, and reassessment; those that remain inconsistent are discarded. The curated corpus contains approximately \textbf{1.0 million trajectories}, totaling \textbf{8,000 hours} of robot interaction data.

\paragraph{Evaluation datasets and splits.}
We evaluate video prediction on the held-out test splits of AgiBotWorld Beta, RoboMIND, and RoboTwin. Each source is partitioned into training, validation, and test splits at a $90\%/5\%/5\%$ ratio before temporal clip extraction. For AgiBotWorld Beta and RoboMIND, we group trajectories by collection session and stratify by task and embodiment. For RoboTwin, we use the randomized configuration with disjoint randomization seeds and initial object configurations across splits. Validation splits are used only for model selection; video-generation metrics are computed on the fixed test splits. These are within-source partitions rather than entirely unseen datasets.

\paragraph{Implementation details.}
Unless otherwise stated, \xsimfull{} processes 93-frame clips at $480\times640$ resolution with frame-aligned action values and action videos, conditioning on the first RGB frame and task instruction to predict the remaining 92 frames. We train for 500K iterations on 32 NVIDIA B200 GPUs with a global batch size of 32 and no gradient accumulation. The learning rate is $3.05\times10^{-5}$, with a 1,000-step linear warmup followed by a constant schedule. Each MoE layer uses one shared expert ($N_s=1$), eight routed experts, and sigmoid top-2 routing with route scale $\kappa_r=2.5$. The routed-branch multiplier $s_r(q)$ increases linearly from $0$ to $1$ over the first 2,000 iterations and remains $1$ thereafter.

\paragraph{Evaluation protocol and metrics.}
On each dataset, all compared methods are evaluated on the same test split using their respective action-conditioning representations. We report PSNR and SSIM~\citep{wang2004ssim} for frame-level reconstruction quality, LPIPS~\citep{zhang2018lpips} for perceptual similarity, FID~\citep{heusel2017fid} for image-distribution quality, and FVD~\citep{unterthiner2018fvd} for spatiotemporal video quality. We further adapt EWMBench~\citep{hu2025ewmbench} to assess motion correctness (DYN, HSD, and nDTW), semantic alignment (BLEU and CLIP), and scene consistency (SceneC), and report Diversity to characterize variation among task-consistent rollouts.

\paragraph{VLM evaluator.}
The world model predicts visual observations; a fine-tuned VLM evaluator converts these predictions into task-conditioned feedback for downstream policy evaluation and ranking, action selection, and policy improvement. Building on VLM-based rollout evaluation~\citep{quevedo2025worldgym,tseng2025scalablepolicy,wu2026oscar} and learned progress estimation~\citep{feng2026procvlm}, we adapt Qwen3-VL-2B-Instruct~\citep{bai2025qwen3vl} with LoRA~\citep{hu2022lora}. Training uses simulator and world-model-generated rollouts from the downstream task domains, with manual annotations of terminal success, task progress, and visual validity, together with same-task ranking supervision. With the pretrained backbone weights frozen, we optimize the LoRA adapters and three prediction heads, whose trainable parameters are denoted by $\phi$, for terminal-success confidence $q_{\phi}$, five-level normalized progress $p_{\phi}$, and visual validity $a_{\phi}$. The evaluator takes the task instruction, a success/failure checklist, and 16 uniformly sampled, temporally ordered RGB frames including both endpoints, and is frozen during downstream use. For action selection, we define the utility $u_{\phi}=a_{\phi}[q_{\phi}+\lambda(1-q_{\phi})p_{\phi}]$, where $\lambda$ weights the progress contribution. Policy-evaluation and reward formulations are detailed in Sections~\ref{sec:policy_evaluation} and~\ref{sec:policy_improvement}.

\subsection{Comparison with State-of-the-Art World Models}
\label{sec:baseline_comparison}

We compare against IRASim~\citep{zhu2024irasim} and Ctrl-World~\citep{guo2026ctrlworld}, which use numerical actions, and the hybrid method EnerVerse-AC~\citep{jiang2025enerverseac} on all three evaluation datasets. On AgiBotWorld Beta, we additionally compare against OSCAR~\citep{wu2026oscar}, which uses visual skeletons, and Wan-Move~\citep{chu2025wanmove}, a motion-controllable video model guided by point trajectories. All evaluated baselines are retrained on the same mixed-domain training corpus.
% TODO: Document how Wan-Move's point-trajectory conditions are constructed
% from the evaluation actions and specify its adaptation settings.

Tables~\ref{tab:world_model_comparison} and~\ref{tab:ewmbench_comparison} summarize video quality and action-conditioned generation on AgiBotWorld Beta, RoboMIND, and RoboTwin. \xsimfull{} achieves the best scores on all five video-quality metrics and the highest adapted EWMBench overall score among the methods evaluated on each dataset. Relative to the strongest baseline for each metric, PSNR improves by $4.636$, $2.080$, and $10.343$ dB, respectively, while the adapted EWMBench overall score increases by $0.566$, $0.390$, and $1.073$. The largest PSNR and overall-score gains occur on RoboTwin, where FVD also decreases from $26.600$ for Ctrl-World to $4.980$. These results indicate improved visual fidelity and aggregate action-conditioned generation performance, although individual adapted EWMBench metrics remain mixed: HSD and nDTW on RoboMIND, and SceneC and Diversity on AgiBotWorld Beta and RoboTwin, remain below the corresponding best baselines.

% Let the main comparisons float independently to avoid partially empty pages.
% Required packages: booktabs, multirow
% Independent float so prose and other tables can fill the surrounding pages.

\begin{table}[tbp]
    \centering
    \normalsize
    \caption{Video-quality comparison on three held-out test splits. For each dataset and metric, the best and second-best results are bold and underlined, respectively.}
    \label{tab:world_model_comparison}
    \setlength{\tabcolsep}{5pt}
    \begin{tabular*}{\linewidth}{@{\extracolsep{\fill}}lccccc@{}}
        \toprule
        \multirow{2}{*}{\textbf{Method}}
        & \multicolumn{2}{c}{\textbf{Computation-based}}
        & \multicolumn{3}{c}{\textbf{Model-based}} \\
        \cmidrule(lr){2-3}\cmidrule(lr){4-6}
        & \textbf{PSNR $\uparrow$} & \textbf{SSIM $\uparrow$}
        & \textbf{LPIPS $\downarrow$} & \textbf{FID $\downarrow$} & \textbf{FVD $\downarrow$} \\
        \midrule
        \multicolumn{6}{l}{\textit{AgiBotWorld Beta}} \\
        IRASim                 & 15.796 & 0.544 & 0.028 & 64.418 & 62.548 \\
        Ctrl-World             & 17.550 & 0.682 & \underline{0.022} & 46.600 & 45.100 \\
        OSCAR                  & 16.644 & 0.608 & 0.023 & 27.805 & 32.241 \\
        Wan-Move               & 12.610 & 0.429 & 0.447 & 88.496 & 95.260 \\
        EnerVerse-AC           & \underline{17.640} & \underline{0.769} & 0.023 & \underline{19.510} & \underline{21.390} \\
        \textbf{\xsimfull{}}   & \textbf{22.276} & \textbf{0.796} & \textbf{0.013} & \textbf{14.640} & \textbf{10.850} \\
        \midrule
        \multicolumn{6}{l}{\textit{RoboMIND}} \\
        IRASim                 & 20.102 & 0.709 & 0.019 & 44.597 & 43.414 \\
        Ctrl-World             & \underline{21.770} & \underline{0.716} & \underline{0.015} & \underline{21.800} & \underline{16.700} \\
        EnerVerse-AC           & 17.930 & 0.680 & 0.293 & 67.350 & 63.690 \\
        \textbf{\xsimfull{}}   & \textbf{23.850} & \textbf{0.835} & \textbf{0.012} & \textbf{17.610} & \textbf{15.930} \\
        \midrule
        \multicolumn{6}{l}{\textit{RoboTwin}} \\
        IRASim                 & 17.121 & 0.666 & 0.025 & 69.777 & 57.350 \\
        Ctrl-World             & \underline{20.040} & \underline{0.804} & \underline{0.016} & \underline{31.400} & \underline{26.600} \\
        EnerVerse-AC           & 19.680 & 0.779 & 0.221 & 32.450 & 29.700 \\
        \textbf{\xsimfull{}}   & \textbf{30.383} & \textbf{0.942} & \textbf{0.006} & \textbf{8.910} & \textbf{4.980} \\
        \bottomrule
    \end{tabular*}
    \par
\end{table}

% Required packages: booktabs, multirow
% Independent float so prose and other tables can fill the surrounding pages.

\begin{table}[tbp]
    \centering
    \normalsize
    \caption{Adapted EWMBench results on the same test splits. Primitive scores are normalized to $[0,1]$; higher is better. Overall denotes the adapted EWMBench overall score. The best and second-best results are bold and underlined, respectively.}
    \label{tab:ewmbench_comparison}
    \setlength{\tabcolsep}{2pt}
    \renewcommand{\arraystretch}{1.03}
    \begin{tabular*}{\linewidth}{@{\extracolsep{\fill}}p{0.24\linewidth}cccccccc@{}}
        \toprule
        \multirow{2}{*}{\textbf{Method}}
        & \multicolumn{3}{c}{\textbf{Motion}}
        & \multicolumn{3}{c}{\textbf{Semantics}}
        & \multicolumn{1}{c}{\textbf{Scene}}
        & \multirow{2}{*}{\shortstack{\textbf{Overall}\\$\uparrow$}} \\
        \cmidrule(lr){2-4}\cmidrule(lr){5-7}\cmidrule(lr){8-8}
        & \shortstack{\textbf{DYN}\\$\uparrow$} & \shortstack{\textbf{HSD}\\$\uparrow$} & \shortstack{\textbf{nDTW}\\$\uparrow$}
        & \shortstack{\textbf{BLEU}\\$\uparrow$} & \shortstack{\textbf{CLIP}\\$\uparrow$}
        & \shortstack{\textbf{Div.}\\$\uparrow$} & \shortstack{\textbf{SceneC}\\$\uparrow$} & \\
        \midrule
        \multicolumn{9}{l}{\textit{AgiBotWorld Beta}} \\
        IRASim & 0.435 & 0.673 & 0.639 & 0.234 & 0.827 & 0.138 & 0.626 & 3.571 \\
        Ctrl-World & \underline{0.556} & 0.904 & 0.889 & 0.300 & 0.848 & 0.197 & \underline{0.707} & 4.401 \\
        OSCAR & 0.488 & 0.823 & 0.811 & \underline{0.400} & \underline{0.904} & 0.192 & 0.685 & 4.303 \\
        Wan-Move & 0.220 & 0.637 & 0.553 & 0.227 & 0.829 & 0.182 & 0.597 & 3.244 \\
        EnerVerse-AC & 0.436 & \underline{0.915} & \underline{0.904} & 0.365 & 0.886 & \textbf{0.202} & \textbf{0.724} & \underline{4.431} \\
        \textbf{\xsimfull{} (Ours)} & \textbf{0.715} & \textbf{0.959} & \textbf{0.962} & \textbf{0.524} & \textbf{0.933} & \underline{0.201} & 0.703 & \textbf{4.997} \\
        \midrule
        \multicolumn{9}{l}{\textit{RoboMIND}} \\
        IRASim & \underline{0.301} & 0.394 & 0.318 & 0.310 & 0.856 & 0.112 & 0.788 & 3.079 \\
        Ctrl-World & 0.214 & \textbf{0.938} & \textbf{0.853} & \underline{0.419} & \underline{0.873} & \underline{0.142} & \underline{0.837} & \underline{4.276} \\
        EnerVerse-AC & 0.170 & 0.419 & 0.393 & 0.257 & 0.839 & 0.127 & 0.776 & 2.982 \\
        \textbf{\xsimfull{} (Ours)} & \textbf{0.591} & \underline{0.838} & \underline{0.810} & \textbf{0.495} & \textbf{0.926} & \textbf{0.157} & \textbf{0.849} & \textbf{4.666} \\
        \midrule
        \multicolumn{9}{l}{\textit{RoboTwin}} \\
        IRASim & \underline{0.395} & 0.612 & 0.551 & 0.277 & 0.785 & 0.122 & 0.640 & 3.383 \\
        Ctrl-World & 0.155 & \underline{0.891} & \underline{0.856} & \underline{0.374} & 0.826 & \textbf{0.163} & \textbf{0.708} & \underline{3.971} \\
        EnerVerse-AC & 0.293 & 0.675 & 0.615 & 0.333 & \underline{0.844} & 0.152 & \underline{0.699} & 3.610 \\
        \textbf{\xsimfull{} (Ours)} & \textbf{0.821} & \textbf{0.972} & \textbf{0.976} & \textbf{0.531} & \textbf{0.918} & \underline{0.160} & 0.666 & \textbf{5.044} \\
        \bottomrule
    \end{tabular*}
    \par
\end{table}

\subsection{Qualitative Generation Results}
\label{sec:qualitative_results}

\begin{figure}[tbp]
    \centering
    % Remove only the top timeline strip; all task labels and frames remain.
    % The caption states the same time direction explicitly.
    \begingroup
    \ifdefined\reportincludegraphics\let\includegraphics\reportincludegraphics\fi
    \includegraphics[width=\linewidth,trim=0 0 0 26bp,clip]{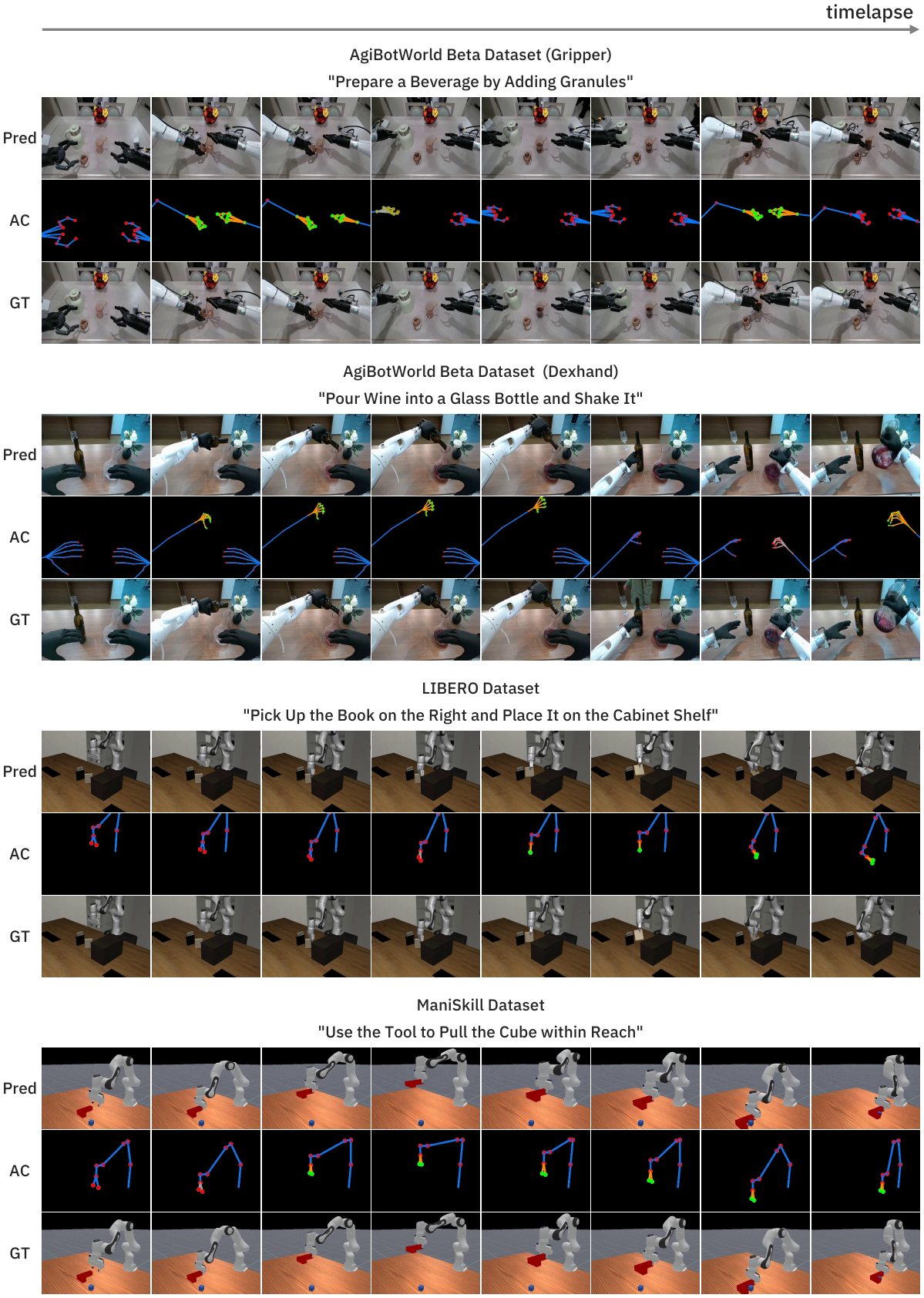}
    \endgroup
    \setlength{\abovecaptionskip}{3pt}
    \caption{\textbf{Multi-source rollouts (Part I).} AgiBotWorld Beta, LIBERO, and ManiSkill. Rows: prediction (Pred), action video (AC), and ground truth (GT); time advances left to right.}
    \label{fig:multisource1}
\end{figure}

\begin{figure}[tbp]
    \centering
    \begingroup
    \ifdefined\reportincludegraphics\let\includegraphics\reportincludegraphics\fi
    \includegraphics[width=\linewidth,trim=0 0 0 26bp,clip]{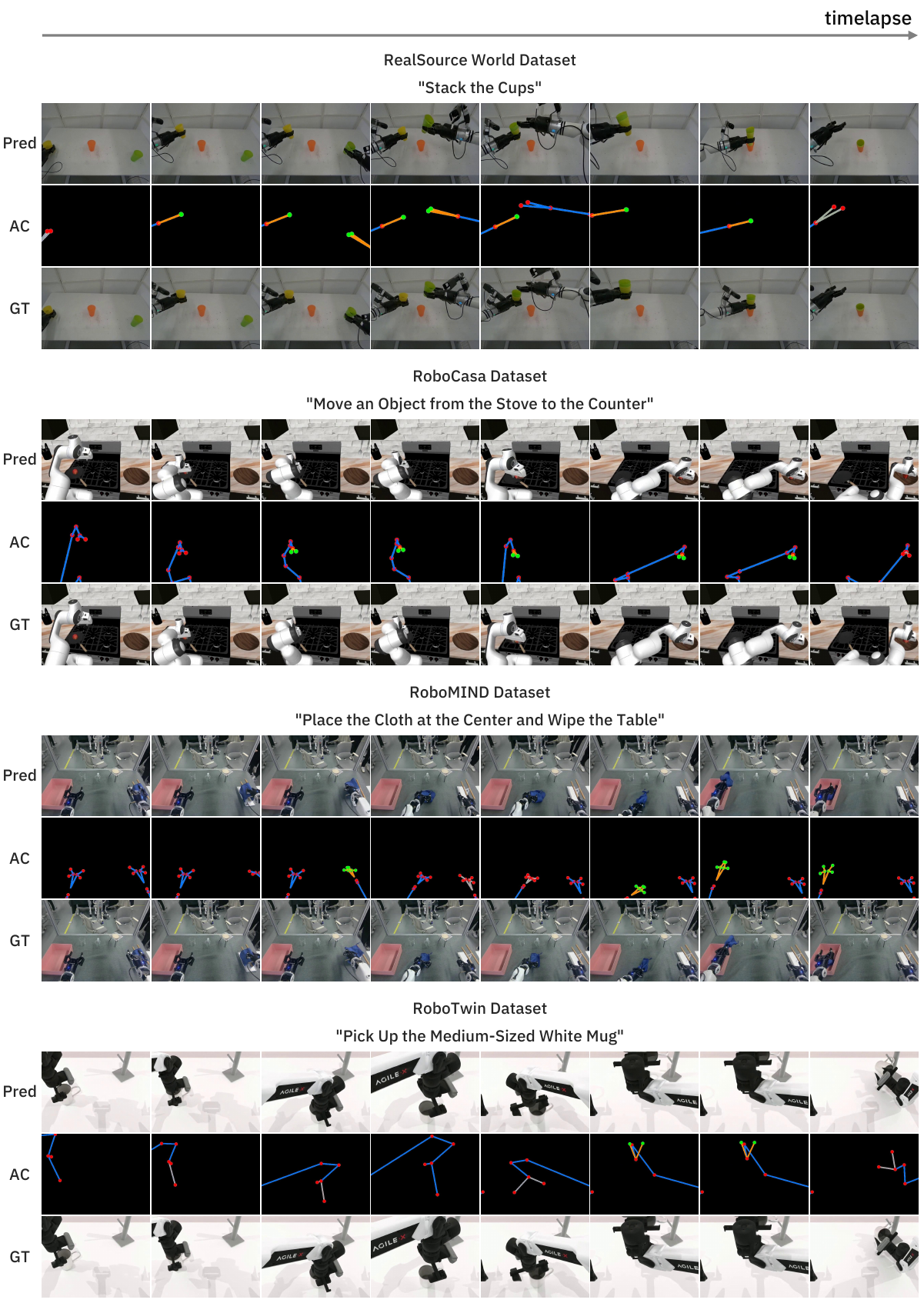}
    \endgroup
    \setlength{\abovecaptionskip}{3pt}
    \caption{\textbf{Multi-source rollouts (Part II).} RealSource World, RoboCasa, RoboMIND, and RoboTwin. Rows: Pred (prediction), AC (action video), GT (ground truth); time runs left to right.}
    \label{fig:multisource2}
\end{figure}

Figures~\ref{fig:multisource1} and~\ref{fig:multisource2} present representative action-conditioned rollouts across real-world and simulated data sources. For each task, the rows show the generated video, the frame-aligned action-video condition, and the ground-truth video, while columns progress from left to right in time. The examples cover single-arm and bimanual systems, parallel grippers and dexterous hands, and diverse object interactions. Across these settings, \xsimfull{} closely follows the commanded whole-arm trajectories and gripper-state transitions while preserving temporal coherence and scene appearance. The predicted object motions are also consistent with the robot actions, indicating effective action control across heterogeneous embodiments and domains.

\subsection{Ablation Studies and Design Analysis}
\label{sec:ablation}

\paragraph{Mixed-domain training and sparse MoE.}
Table~\ref{tab:data_moe_ablation} presents a cumulative ablation on AgiBotWorld Beta, comparing a dense baseline trained on real data, the same dense model trained on mixed-domain data, and an MoE variant trained on mixed-domain data. All variants share the action conditions, injection architecture, optimization schedule, and training iterations. Adding simulated trajectories improves PSNR by $0.548$ dB and reduces FVD by $3.030$; replacing the dense feed-forward layers with sparse MoE layers under mixed-domain training yields further gains of $1.214$ dB and $6.530$, respectively. These results support the two successive design choices, while the mixed-domain improvement reflects changes in both data diversity and corpus size rather than an isolated effect of domain composition.

% Required packages: booktabs, graphicx

\begin{table}[tbp]
    \centering
    \normalsize
    \caption{Cumulative ablation of mixed-domain training and sparse MoE on the held-out test split of AgiBotWorld Beta. Dense and Sparse MoE denote the backbone variants with dense feed-forward layers and sparse MoE layers, respectively. All variants use the same action conditions, injection architecture, optimization schedule, and number of training iterations. The best results are highlighted in bold.}
    \label{tab:data_moe_ablation}
    \setlength{\tabcolsep}{3pt}
    \begin{tabular*}{\linewidth}{@{\extracolsep{\fill}}p{0.34\linewidth}ccccc@{}}
        \toprule
        \textbf{Training setting}
        & \textbf{PSNR $\uparrow$}
        & \textbf{SSIM $\uparrow$}
        & \textbf{LPIPS $\downarrow$}
        & \textbf{FID $\downarrow$}
        & \textbf{FVD $\downarrow$} \\
        \midrule
        Real only (Dense)                  & 20.514 & 0.7512 & 0.0186 & 23.870 & 20.410 \\
        Real + Simulation (Dense)          & 21.062 & 0.7668 & 0.0169 & 20.620 & 17.380 \\
        Real + Simulation + Sparse MoE     & \textbf{22.276} & \textbf{0.7960} & \textbf{0.0125} & \textbf{14.640} & \textbf{10.850} \\
        \bottomrule
    \end{tabular*}
\end{table}

\paragraph{Contribution of complementary action conditions.}
Table~\ref{tab:action_condition_ablation} compares four action-conditioning configurations on AgiBotWorld Beta and RoboMIND. The no-action baseline evaluates the video generation model without numerical or visual action inputs. Both single-condition variants improve over this baseline, with action videos outperforming numerical actions on all reported metrics in this ablation. Combining both conditions yields the best prediction quality, increasing PSNR over action video alone by $1.000$ and $0.748$ dB on the two datasets, respectively, with corresponding LPIPS reductions from $0.0158$ to $0.0125$ and from $0.0132$ to $0.0115$. These results support combining explicit image-space motion guidance with numerical configuration information, rather than replacing numerical actions with visual conditions alone.

% Required packages: booktabs, graphicx, amssymb

\begin{table}[tbp]
    \centering
    \normalsize
    \caption{Ablation of the complementary action conditions on the held-out test splits of AgiBotWorld Beta and RoboMIND. The no-action baseline evaluates video generation without numerical or visual action inputs. All variants use the same backbone, training data, and evaluation protocol. The best results are highlighted in bold.}
    \label{tab:action_condition_ablation}
    \setlength{\tabcolsep}{2pt}
    \begin{tabular*}{\linewidth}{@{\extracolsep{\fill}}p{0.19\linewidth}cccccccc@{}}
        \toprule
        \multirow{2}{*}{\textbf{Setting}}
        & \multirow{2}{*}{\shortstack{\textbf{Action}\\\textbf{value}}}
        & \multirow{2}{*}{\shortstack{\textbf{Action}\\\textbf{video}}}
        & \multicolumn{3}{c}{\textbf{AgiBotWorld Beta}}
        & \multicolumn{3}{c}{\textbf{RoboMIND}} \\
        \cmidrule(lr){4-6}\cmidrule(lr){7-9}
        & &
        & \shortstack{\textbf{PSNR}\\$\uparrow$}
        & \shortstack{\textbf{SSIM}\\$\uparrow$}
        & \shortstack{\textbf{LPIPS}\\$\downarrow$}
        & \shortstack{\textbf{PSNR}\\$\uparrow$}
        & \shortstack{\textbf{SSIM}\\$\uparrow$}
        & \shortstack{\textbf{LPIPS}\\$\downarrow$} \\
        \midrule
        \shortstack[l]{No action\\condition}
        & $\times$ & $\times$ & 17.684 & 0.6513 & 0.0287 & 19.640 & 0.7195 & 0.0248 \\
        Action value only
        & $\checkmark$ & $\times$ & 19.238 & 0.7126 & 0.0214 & 21.356 & 0.7768 & 0.0176 \\
        Action video only
        & $\times$ & $\checkmark$ & 21.276 & 0.7668 & 0.0158 & 23.102 & 0.8209 & 0.0132 \\
        Full dual condition
        & $\checkmark$ & $\checkmark$ & \textbf{22.276} & \textbf{0.7960} & \textbf{0.0125} & \textbf{23.850} & \textbf{0.8350} & \textbf{0.0115} \\
        \bottomrule
    \end{tabular*}
\end{table}

\paragraph{Injection architecture.}
Table~\ref{tab:injection_architecture_ablation} compares token concatenation, late fusion, two opposite interleaving orders, and dual injection into every block of the Video DiT. Both interleaved variants outperform token concatenation and late fusion, with similar scores across the reported metrics. Dual injection achieves the highest scores but improves PSNR over the selected interleaving by only $0.075$ dB on AgiBotWorld Beta and $0.051$ dB on RoboMIND.
This suggests that repeatedly injecting both conditions at every block provides limited additional benefit. Since numerical and visual actions encode complementary aspects of the same trajectory, their information can instead be progressively integrated through the shared hidden representations across network depth. We therefore adopt interleaved injection, which alternates the two conditioning pathways across blocks of the same Video DiT and achieves nearly the same prediction quality as dual injection with fewer conditioning modules.
% It also requires both conditioning branches in every block, whereas interleaving uses one branch per block. We therefore adopt interleaved injection, which achieves comparable prediction quality with fewer conditioning modules.

% Required packages: booktabs, graphicx

\begin{table}[tbp]
    \centering
    \normalsize
    \caption{Ablation of action-condition injection architectures on the held-out test splits of AgiBotWorld Beta and RoboMIND. Dual injection instantiates both conditioning branches at every block, whereas interleaved variants alternate the branches across blocks. The best results are highlighted in bold, and the selected configuration is identified by its bold method name.}
    \label{tab:injection_architecture_ablation}
    \setlength{\tabcolsep}{3pt}
    \begin{tabular*}{\linewidth}{@{\extracolsep{\fill}}p{0.25\linewidth}cccccc@{}}
        \toprule
        \multirow{2}{*}{\shortstack[l]{\textbf{Injection}\\\textbf{architecture}}}
        & \multicolumn{3}{c}{\textbf{AgiBotWorld Beta}}
        & \multicolumn{3}{c}{\textbf{RoboMIND}} \\
        \cmidrule(lr){2-4}\cmidrule(lr){5-7}
        & \shortstack{\textbf{PSNR}\\$\uparrow$}
        & \shortstack{\textbf{SSIM}\\$\uparrow$}
        & \shortstack{\textbf{LPIPS}\\$\downarrow$}
        & \shortstack{\textbf{PSNR}\\$\uparrow$}
        & \shortstack{\textbf{SSIM}\\$\uparrow$}
        & \shortstack{\textbf{LPIPS}\\$\downarrow$} \\
        \midrule
        Token concat.
        & 21.214 & 0.7601 & 0.0164 & 21.903 & 0.7815 & 0.0170 \\
        Late fusion
        & 21.372 & 0.7664 & 0.0155 & 22.931 & 0.8108 & 0.0138 \\
        Reverse interleave
        & 22.241 & 0.7953 & 0.0126 & 23.821 & 0.8344 & 0.0116 \\
        \textbf{Interleaved (ours)}
        & 22.276 & 0.7960 & 0.0125 & 23.850 & 0.8350 & 0.0115 \\
        Dual every block
        & \textbf{22.351} & \textbf{0.7976} & \textbf{0.0123} & \textbf{23.901} & \textbf{0.8362} & \textbf{0.0114} \\
        \bottomrule
    \end{tabular*}
\end{table}

\subsection{Data Generation}
\label{sec:data_engine}

% We assess \xsimfull{} as a data-generation engine across three progressively demanding settings: generation efficiency -- Can \xsimfull{} generate data efficiently?, cross-domain data construction -- Is generated data useful for policy training?, and policy learning with limited demonstrations -- Does it help when demonstrations are scarce?. 
% We assess \xsimfull{} as a data-generation engine across three progressively demanding settings: generation efficiency, cross-domain data construction, and policy learning with limited demonstrations.
We assess \xsimfull{} as a data-generation engine across three progressively demanding settings: \textbf{generation efficiency}, examining whether trajectories can be generated efficiently; \textbf{cross-domain data construction}, evaluating whether generated rollouts provide useful supervision for policy training; and \textbf{policy learning with limited demonstrations}, assessing whether augmentation remains beneficial when collected data are scarce.
In both policy studies, trajectory counts are specified per task. Within each study, compared policies share a fixed set of 200 previously unseen evaluation seeds per task under the RoboTwin randomized configuration, disjoint from training-data construction.

\paragraph{Generation efficiency.}
Following robot data-generation evaluations~\citep{he2026vdreamer}, we compare the accelerated four-step \xsimfull{} model, RoboTwin simulation, and manual collection for complete task trajectories, rather than individual video clips, at a matched trajectory horizon. Timing includes inference and video serialization for \xsimfull{}, stepping, rendering, and serialization for RoboTwin, and physical execution and recording for manual collection. Nominal throughput is derived from these times, excluding manual reset and upstream preparation such as model training, simulator construction, and initial-frame editing. We separately report reset time for manual scene reset and preparation between episodes, excluding teleoperation during execution.

% Required packages: booktabs, graphicx

\begin{table}[tbp]
    \centering
    \normalsize
    \caption{Data generation efficiency of \xsimfull{} compared with manual collection and RoboTwin simulation for complete trajectories at matched horizons. The accelerated four-step causal model is used in both comparisons. Times cover generation or execution and recording; nominal throughput excludes upstream preparation and the separately reported manual reset time. Zero reset time denotes no manual intervention between episodes.}
    \label{tab:data_engine_efficiency}
    \setlength{\tabcolsep}{4pt}
    \begin{tabular*}{\linewidth}{@{\extracolsep{\fill}}p{0.25\linewidth}p{0.22\linewidth}ccc@{}}
        \toprule
        \textbf{Data source} & \textbf{Resource} & \shortstack{\textbf{Sec./traj.}\\$\downarrow$} & \shortstack{\textbf{Traj./hour}\\$\uparrow$} & \shortstack{\textbf{Reset time}\\\textbf{(min./traj.)} $\downarrow$} \\
        \midrule
        \multicolumn{5}{l}{\textit{Real-world comparison}} \\
        Manual collection   & Robot + operator & 60  & 60   & 0.4 \\
        \xsimfull{}          & GPU              & 24  & 150  & 0 \\
        \midrule
        \multicolumn{5}{l}{\textit{Simulation comparison}} \\
        RoboTwin simulator  & Simulator + GPU  & 55  & 65.5 & 0 \\
        \xsimfull{}          & GPU              & 19  & 189.5 & 0 \\
        \bottomrule
    \end{tabular*}
\end{table}

Table~\ref{tab:data_engine_efficiency} reports generation efficiency separately for the real-world and simulation comparisons. In the real-world comparison, \xsimfull{} generates a complete task trajectory in $24$ seconds, compared with $60$ seconds for manual collection, yielding nominal throughputs of $150$ and $60$ trajectories per hour, respectively ($2.50\times$). In the simulation comparison, \xsimfull{} requires $19$ seconds per trajectory, compared with $55$ seconds for RoboTwin, yielding approximately $189.5$ and $65.5$ trajectories per hour, respectively ($2.89\times$). These throughput ratios exclude manual reset time. Manual collection additionally requires $0.4$ minutes of reset time per trajectory, whereas neither automated method requires manual reset between episodes.

\paragraph{Cross-domain augmentation.}
We compare two data-construction strategies through downstream policy performance: augmenting clean demonstrations with \xsimfull{} and collecting randomized demonstrations directly in RoboTwin. Starting from 50 trajectories per task in the non-randomized \texttt{demo\_clean} configuration~\citep{chen2025robotwin2}, we use GPT-Image-2~\citep{openai2026gptimage2} to produce ten initial-frame variants resembling \texttt{demo\_randomized} scenes. \xsimfull{} generates a rollout from each variant using the original actions. We train a $\pi_{0.5}$ VLA policy~\citep{physicalintelligence2025pi05} on the 50 clean and 500 generated trajectories, and compare it with an identically configured policy trained on 50 directly collected randomized trajectories per task.

% Required package: booktabs

\begin{table}[tbp]
    \centering
    \normalsize
    \caption{Cross-domain data augmentation for downstream $\pi_{0.5}$ policy learning. Per task, \xsimfull{} augmentation combines 50 clean demonstrations with 500 generated rollouts from edited initial observations; direct collection uses 50 randomized simulator trajectories. Identically configured policies are evaluated under RoboTwin's randomized configuration, with success averaged over five tasks and 200 unseen seeds per task.}
    \label{tab:data_engine_quality}
    \setlength{\tabcolsep}{5pt}
    \begin{tabular*}{\linewidth}{@{\extracolsep{\fill}}p{0.30\linewidth}cccc@{}}
        \toprule
        \textbf{Training data} & \shortstack{\textbf{Seed}\\\textbf{traj.}} & \shortstack{\textbf{Aug.}\\\textbf{/ seed}} & \shortstack{\textbf{Total}\\\textbf{traj.}} & \shortstack{\textbf{Success}\\\textbf{(\%)}} \\
        \midrule
        RoboTwin randomized         & 50 & 0  & 50  & 70.0 \\
        \xsimfull{} augmentation    & 50 & 10 & 550 & 72.0 \\
        \bottomrule
    \end{tabular*}
\end{table}

Table~\ref{tab:data_engine_quality} reports mean success rates of $72.0\%$ for \xsimfull{} augmentation and $70.0\%$ for direct simulator collection across five tasks. The similar observed performance supports \xsimfull{}-based data construction as an alternative for downstream policy training.

\paragraph{Scaling with limited demonstrations.}
We next evaluate augmentation with 10, 30, or 50 original randomized trajectories per task. Using the same strategy as in the cross-domain study, we generate ten additional rollouts per original trajectory. We compare $\pi_{0.5}$ policies trained on the originals alone with those trained on augmented sets of 110, 330, and 550 trajectories per task.

% Required package: booktabs

\begin{table}[tbp]
    \centering
    \normalsize
    \caption{Effect of \xsimfull{} augmentation on $\pi_{0.5}$ policy success at three demonstration budgets. Original-only training is compared with adding ten generated rollouts per original trajectory. Counts are per task; evaluation uses RoboTwin's randomized configuration with 200 unseen seeds per task. Parenthesized values report percentage-point gains over the corresponding original-only baseline.}
    \label{tab:data_engine_scaling}
    \setlength{\tabcolsep}{5pt}
    \begin{tabular*}{\linewidth}{@{\extracolsep{\fill}}p{0.30\linewidth}cccc@{}}
        \toprule
        \textbf{Setting} & \shortstack{\textbf{Original}\\\textbf{traj.}} & \shortstack{\textbf{Generated}\\\textbf{traj.}} & \shortstack{\textbf{Total}\\\textbf{traj.}} & \shortstack{\textbf{Success}\\\textbf{(\%)}} \\
        \midrule
        Original only                 & 10 & 0   & 10  & 28.5 \\
        Original + \xsimfull{}        & 10 & 100 & 110 & 64.5 ($\uparrow 36.0$) \\
        Original only                 & 30 & 0   & 30  & 57.0 \\
        Original + \xsimfull{}        & 30 & 300 & 330 & 87.0 ($\uparrow 30.0$) \\
        Original only                 & 50 & 0   & 50  & 70.0 \\
        Original + \xsimfull{}        & 50 & 500 & 550 & 93.0 ($\uparrow 23.0$) \\
        \bottomrule
    \end{tabular*}
\end{table}

Table~\ref{tab:data_engine_scaling} shows success-rate gains of $36.0$, $30.0$, and $23.0$ percentage points at the three demonstration budgets, respectively. Augmentation with only 10 original trajectories per task reaches $64.5\%$, compared with $70.0\%$ for 50 originals without augmentation. These results indicate that generated rollouts can reduce the need for additional collected demonstrations in the evaluated setting, with the largest absolute gain at the smallest demonstration budget.

\subsection{Policy Evaluation and Ranking}
\label{sec:policy_evaluation}

Following world-model-based policy evaluation~\citep{quevedo2025worldgym,tseng2025scalablepolicy,jeon2026roboworld}, we next examine whether a pipeline combining \xsimfull{} with the fine-tuned VLM evaluator reproduces policy success rates and rankings measured in RoboTwin. The world model predicts visual outcomes for supplied action trajectories, and the evaluator judges task completion from the generated videos. To assess evaluation robustness across policy performance levels, we select five checkpoints $\{\pi_k\}_{k=1}^{5}$ at different stages of the same VLA training run, representing policies with varying task success rates. These policies provide trajectories for both validation of the frozen VLM evaluator from Section~\ref{sec:experimental_setup} and assessment of the complete world-model--VLM pipeline.

\paragraph{Validation of the VLM evaluator.}
For each policy, we obtain paired RoboTwin and \xsimfull{} rollouts on held-out initial conditions using the same action sequences and initial observations. The validation set contains 1,000 simulator rollouts across five tasks, with binary task-success labels provided by the RoboTwin task checker and human-annotated progress references. Three annotators label a score-stratified subset of 700 corresponding generated videos, with majority vote determining task success and visual validity. Progress labels for simulator and generated videos follow the same annotation protocol: annotators rate how fully the task requirements are satisfied at the end of each video on a five-level scale, and the median rating is normalized to $[0,1]$ as the reference score. Each video is assessed on its own depicted outcome; shared annotation criteria do not imply identical labels for paired rollouts. We report accuracy, balanced accuracy, F1, AUROC, and expected calibration error (ECE) for task-success prediction; MAE and Spearman correlation for progress estimation; and AUROC for visual validity.

Table~\ref{tab:vlm_judge_validation} separately reports evaluator performance on simulator and generated videos. The evaluator achieves terminal accuracies of $92.6\%$ and $91.4\%$, respectively, with a progress MAE of $0.078$ and validity AUROC of $0.953$ on generated videos. These results assess the VLM's ability to judge depicted outcomes, separately from the world model's ability to reproduce simulator outcomes.

% Required packages: booktabs, graphicx

\begin{table}[tbp]
    \centering
    \normalsize
    \caption{Validation of the fine-tuned Qwen3-VL-2B-Instruct evaluator on simulator and generated videos. Upper: terminal-success prediction; lower: progress estimation and visual validity. Success references use the RoboTwin task checker or human majority vote for generated videos; progress and validity use human medians and majority vote, respectively. Threshold-dependent metrics use disjoint calibration data. Each video is judged on its depicted outcome, independently of world-model prediction accuracy.}
    \label{tab:vlm_judge_validation}
    \setlength{\tabcolsep}{3pt}
    \begin{tabular*}{\linewidth}{@{\extracolsep{\fill}}p{0.25\linewidth}rccccc@{}}
        \toprule
        \textbf{Evaluation input}
        & $\boldsymbol{N}$
        & \shortstack{\textbf{Accuracy}\\\textbf{(\%)} $\uparrow$}
        & \shortstack{\textbf{Balanced}\\\textbf{acc. (\%)} $\uparrow$}
        & \shortstack{\textbf{F1 (\%)}\\$\uparrow$}
        & \shortstack{\textbf{AUROC}\\$\uparrow$}
        & \shortstack{\textbf{ECE}\\$\downarrow$} \\
        \midrule
        RoboTwin videos       & 1,000 & 92.6 & 91.8 & 92.1 & 0.964 & 0.041 \\
        \xsimfull{} videos    & 700   & 91.4 & 90.8 & 91.0 & 0.951 & 0.048 \\
        \bottomrule
    \end{tabular*}

    \vspace{4pt}
    \begin{tabular*}{\linewidth}{@{\extracolsep{\fill}}p{0.25\linewidth}rccc@{}}
        \toprule
        \textbf{Evaluation input}
        & $\boldsymbol{N}$
        & \shortstack{\textbf{Progress}\\\textbf{MAE} $\downarrow$}
        & \shortstack{\textbf{Progress}\\$\rho\;\uparrow$}
        & \shortstack{\textbf{Validity}\\\textbf{AUROC} $\uparrow$} \\
        \midrule
        RoboTwin videos      & 1,000 & 0.061 & 0.926 & --    \\
        \xsimfull{} videos   & 700   & 0.078 & 0.901 & 0.953 \\
        \bottomrule
    \end{tabular*}
\end{table}

\paragraph{Policy evaluation and ranking protocol.}
Each policy is evaluated on the same $N=200$ held-out RoboTwin initial conditions. For each condition, the same policy-predicted action trajectory is executed in RoboTwin and supplied to \xsimfull{}. Success is determined by the RoboTwin checker for simulator trials and by the frozen VLM for generated trials. This matched-action protocol evaluates outcome prediction for fixed trajectories, rather than closed-loop policy execution under generated observations. We vary the world-model adaptation budget over 0, 100, 200, 500, and 1,000 task-specific RoboTwin rollouts, with nonzero budgets balanced across the five policies.

For policy $\pi_k$, the estimated success rate is $\mathrm{SR}^{\mathrm{wm}}_k=N^{-1}\sum_{i=1}^{N}z_{\phi}(\widehat{\mathbf{V}}_{k,i},\mathbf{y}_i)$, where $\widehat{\mathbf{V}}_{k,i}$ is the generated rollout including the initial frame for evaluation condition $i$, and $\mathbf{y}_i$ is its instruction. The binary decision is $z_{\phi}=\mathbb{I}[q_{\phi}\geq\tau_{\mathrm{succ}}]$, with $\tau_{\mathrm{succ}}$ denoting the terminal-success threshold. We compare the estimate with the corresponding RoboTwin success rate using Pearson $r$ for linear correlation, Spearman $\rho$ and Kendall $\tau$ for ranking consistency, mean maximum rank violation (MMRV) for ordering errors, and MAE in percentage points for absolute success-rate error~\citep{tseng2025scalablepolicy}.
% TODO: Report the evaluator's success threshold tau_succ and utility weight lambda.

\paragraph{Policy evaluation and ranking results.}
Table~\ref{tab:policy_evaluation_scaling} summarizes agreement across the five adaptation budgets. With 200 world-model adaptation rollouts, the world-model--VLM pipeline reproduces the RoboTwin ranking of all five evaluated checkpoints, achieving Spearman $\rho=1.0$, Kendall $\tau=1.0$, and MMRV $=0$. Larger budgets preserve this ranking agreement while improving absolute success-rate estimates: MAE decreases from $4.0$ percentage points at 200 rollouts to $2.9$ at 500 and $2.4$ at 1,000. We use the 500-rollout variant in subsequent experiments, balancing estimation accuracy against adaptation data requirements; doubling this budget yields a further MAE reduction of $0.5$ points. These results distinguish accurate policy ranking from accurate success-rate estimation within the evaluated checkpoint set.

% Retained as the policy-evaluation table entry point.
% Required package: booktabs
\begin{table}[tbp]
    \centering
    \normalsize
    \caption{Policy evaluation and ranking with the \xsimfull{}--VLM pipeline across five world-model adaptation budgets. Agreement with RoboTwin is measured for five policy checkpoints using matched action trajectories on 200 held-out initial conditions each, with the VLM frozen. MAE reports success-rate error in percentage points; MMRV measures ordering errors using success rates in $[0,1]$, with zero indicating no ranking violation.}
    \label{tab:policy_evaluation_scaling}
    \begin{tabular*}{\linewidth}{@{\extracolsep{\fill}}rccccc@{}}
        \toprule
        \shortstack{\textbf{Adaptation}\\\textbf{rollouts}}
        & \shortstack{\textbf{Pearson}\\$r\,\uparrow$}
        & \shortstack{\textbf{Spearman}\\$\rho\,\uparrow$}
        & \shortstack{\textbf{Kendall}\\$\tau\,\uparrow$}
        & \shortstack{\textbf{MMRV}\\$\downarrow$}
        & \shortstack{\textbf{MAE (pp)}\\$\downarrow$} \\
        \midrule
        0     & 0.9020 & 0.800 & 0.600 & 0.162 & 8.6 \\
        100   & 0.9580 & 0.900 & 0.800 & 0.074 & 5.3 \\
        200   & 0.9760 & \textbf{1.000} & \textbf{1.000} & \textbf{0.000} & 4.0 \\
        500 & 0.9890 & \textbf{1.000} & \textbf{1.000} & \textbf{0.000} & 2.9 \\
        1,000 & \textbf{0.9940} & \textbf{1.000} & \textbf{1.000} & \textbf{0.000} & \textbf{2.4} \\
        \bottomrule
    \end{tabular*}
\end{table}

\subsection{Action Selection}
\label{sec:action_selection}

Beyond policy-level evaluation, we test whether VLM-guided selection among outcomes predicted by \xsimfull{} improves task success. A frozen policy proposes candidate action sequences, the world model predicts their outcomes, and the selector chooses a candidate using the VLM-derived utility.

\paragraph{Candidate generation and selection.}
Given an initial observation $\mathbf{x}_0$ and instruction $\mathbf{y}$, Diffusion Policy~\citep{chi2023diffusion} samples $M=10$ action sequences over a common horizon using independent diffusion noise. Following Best-of-$N$ selection~\citep{alzayer2026masked}, the accelerated 500-rollout adapted \xsimfull{} model predicts each candidate's outcome, and the frozen VLM evaluator ranks the resulting videos using $u_{\phi}$ from Section~\ref{sec:experimental_setup}:
\begin{equation}
    m^{*}
    =\underset{m\in\{1,\ldots,M\}}{\operatorname{argmax}}
      \;u_{\phi}\!\left(
        \widehat{\mathbf{V}}^{(m)},\mathbf{y}
      \right),
    \qquad
    \widehat{\mathbf{X}}^{\mathrm{tar},(m)}
    \sim p_{\theta}\!\left(
        \cdot\mid\mathbf{x}_{0},\mathbf{A}^{(m)},\mathbf{y}
    \right).
    \label{eq:action_selection}
\end{equation}
Here, $\mathbf{A}^{(m)}$ is candidate $m$ in the unified configuration layout, $\widehat{\mathbf{X}}^{\mathrm{tar},(m)}$ is its predicted future video, and $\widehat{\mathbf{V}}^{(m)}=[\mathbf{x}_0;\widehat{\mathbf{X}}^{\mathrm{tar},(m)}]$ is the complete video scored by the evaluator. The selected index is $m^*$, and only $\mathbf{A}^{(m^{*})}$ is executed in RoboTwin, open-loop without replanning.

\paragraph{Evaluation protocol and baselines.}
We evaluate five RoboTwin tasks using 200 held-out initial conditions per task, disjoint from world-model adaptation and VLM-evaluator calibration. All methods share candidate pools. Baselines include \emph{Single sample} (the first candidate), \emph{Random-of-10} (uniform selection), \emph{VLM-direct} (scoring the initial observation and rendered action condition without future prediction), and \emph{Ctrl-World selection}~\citep{guo2026ctrlworld} (replacing \xsimfull{} as the rollout generator). The two world-model selectors share the adaptation budget, prediction horizon, output resolution, and VLM evaluator. \emph{Oracle Best-of-10} provides a candidate-pool upper bound using task-checker outcomes from executions of all candidates in cloned simulator states; these outcomes are unavailable to the learned selectors.
% TODO: Confirm whether tasks, in addition to initial conditions, are held out;
% clarify the source
% of adaptation trajectories and verify matched 500-rollout baseline adaptation.
% TODO: Specify how VLM-direct accepts the rendered action condition and whether
% it uses the same evaluator weights or an input-specific adaptation.

We report task-checker success rate, improvement over Single sample, and oracle gap, with differences in percentage points. Any-success@$M$ is the fraction of pools containing a successful candidate, while selection recall is the success rate conditional on such a pool.
% TODO: Report paired bootstrap confidence intervals, including
% the resampling unit and aggregation across tasks.

% Required packages: booktabs, graphicx, amssymb

\begin{table}[tbp]
    \centering
    \normalsize
    \caption{Action selection results averaged over five RoboTwin tasks. All methods use the same Diffusion Policy candidate pools and 200 held-out initial conditions per task. Ctrl-World and \xsimfull{} use the same adaptation budget and VLM evaluator; Oracle Best-of-10 uses simulator outcomes only as an upper bound.}
    \label{tab:action_selection}
    \setlength{\tabcolsep}{3pt}
    \begin{tabular*}{\linewidth}{@{\extracolsep{\fill}}p{0.27\linewidth}ccccc@{}}
        \toprule
        \textbf{Method}
        & \shortstack{\textbf{Predicted}\\\textbf{future}}
        & \shortstack{\textbf{Success}\\\textbf{(\%)} $\uparrow$}
        & \shortstack{$\boldsymbol{\Delta}$\textbf{SR}\\\textbf{(pp)} $\uparrow$}
        & \shortstack{\textbf{Oracle gap}\\\textbf{(pp)} $\downarrow$}
        & \shortstack{\textbf{Selection}\\\textbf{recall (\%)} $\uparrow$} \\
        \midrule
        Single sample                & --          & 43.2 & $0.0$  & 24.9 & 63.4 \\
        Random-of-10                 & --          & 43.5 & $+0.3$ & 24.6 & 63.9 \\
        VLM-direct                   & --          & 48.7 & $+5.5$ & 19.4 & 71.5 \\
        Ctrl-World selection         & \checkmark  & \underline{52.3} & $\underline{+9.1}$ & \underline{15.8} & \underline{76.8} \\
        \xsimfull{} selection        & \checkmark  & \textbf{63.8} & $\mathbf{+20.6}$ & \textbf{4.3} & \textbf{93.7} \\
        Oracle Best-of-10            & RoboTwin    & 68.1 & $+24.9$ & 0.0 & 100.0 \\
        \bottomrule
    \end{tabular*}
\end{table}

Table~\ref{tab:action_selection} shows that the selector combining \xsimfull{} with the frozen VLM evaluator achieves $63.8\%$ success, exceeding Single sample and Ctrl-World selection by $20.6$ and $11.5$ percentage points. Its $93.7\%$ selection recall leaves a $4.3$-point gap to the oracle. The $15.1$-point gain over VLM-direct supports the utility of predicted visual outcomes for selection, whereas Random-of-10 performs similarly to Single sample.

\paragraph{Test-time scaling.}
We vary $M\in\{1,2,4,8,10\}$ using nested candidate pools and measure world-model generation and VLM scoring latency on four GPUs. Table~\ref{tab:action_selection_scaling} shows that increasing $M$ from 1 to 10 improves success from $43.2\%$ to $63.8\%$, but recall falls from $100\%$ to $93.7\%$ as selection misses some available successful candidates; recall at $M=1$ is $100\%$ by construction. Increasing $M$ from 8 to 10 yields only $1.3$ additional percentage points of success while latency rises from $10.7$ to $15.9$ seconds, highlighting the success--latency trade-off.
% TODO: Specify GPU model, batching, trajectory horizon, and whether candidate
% sampling and action-condition rendering are included in decision latency.

% Required package: booktabs

\begin{table}[tbp]
    \centering
    \normalsize
    \caption{Test-time scaling of \xsimfull{} action selection with the number of Diffusion Policy candidates. Candidate sets are nested across $M$. Latency covers world-model generation and VLM scoring under a fixed four-GPU setup. Any-success is the oracle success rate within each candidate pool; selection recall is conditional on the pool containing a successful candidate.}
    \label{tab:action_selection_scaling}
    \setlength{\tabcolsep}{5pt}
    \begin{tabular*}{\linewidth}{@{\extracolsep{\fill}}rcccc@{}}
        \toprule
        $\boldsymbol{M}$
        & \shortstack{\textbf{Any-success}\\\textbf{(\%)} $\uparrow$}
        & \shortstack{\textbf{Selected success}\\\textbf{(\%)} $\uparrow$}
        & \shortstack{\textbf{Selection recall}\\\textbf{(\%)} $\uparrow$}
        & \shortstack{\textbf{Sec./decision}\\$\downarrow$} \\
        \midrule
        1  & 43.2 & 43.2 & 100.0 & 5.2  \\
        2  & 52.6 & 51.2 & 97.3  & 5.3  \\
        4  & 60.4 & 58.0 & 96.0  & 5.5  \\
        8  & 66.4 & 62.5 & 94.1  & 10.7 \\
        10 & \textbf{68.1} & \textbf{63.8} & 93.7 & 15.9 \\
        \bottomrule
    \end{tabular*}
\end{table}

\subsection{Policy Improvement}
\label{sec:policy_improvement}

We finally evaluate policy improvement through closed-loop rollouts from \xsimfull{} and rewards derived from the fine-tuned VLM evaluator. The policy generates actions from predicted observations, the evaluator supplies task-related scores, and GRPO uses returns constructed from these scores to update the policy. Following World-Gymnast~\citep{sharma2026worldgymnast}, we initialize from a task-specific OpenVLA-OFT checkpoint~\citep{kim2025openvlaoft,kim2024openvla} and use a Llama-2 language-model head in place of the L1 regression head to obtain action-token probabilities for GRPO, while retaining action-chunk prediction. Policy optimization uses the accelerated four-step causal version of the 500-rollout adapted model from Section~\ref{sec:policy_evaluation}. The world model and VLM evaluator remain frozen.
% TODO: Specify how the token-based action head is initialized and supervised
% before RL, and confirm the shared initialization across policy baselines.

\paragraph{Policy-to-world-model action interface.}
The policy predicts fixed-length chunks of joint-angle targets and gripper or hand states. We decode and denormalize the actions, then pack them into the 28-dimensional layout with zero-filled unused entries (Section~\ref{sec:action_representation}). Each chunk is interpolated to 21 configurations, including the current configuration at the conditioning frame, and rendered through the URDF pipeline in Figure~\ref{fig:action_video_rendering}. This produces synchronized numerical and visual action conditions. The final commanded configuration initializes the next transition; no privileged object or simulator state is provided to the policy.

\paragraph{Closed-loop world-model rollouts.}
Let $\vartheta$ denote the trainable policy parameters, distinct from world-model parameters $\theta$ and evaluator parameters $\phi$, and let $\vartheta_{\mathrm{old}}$ denote the behavior-policy parameters used to collect a rollout group. For each instruction $\mathbf y$ and initial observation $\mathbf x_0$, we sample $K$ imagined trajectories over $H$ decision steps. At decision index $h\in\{0,\ldots,H-1\}$ in trajectory $k\in\{1,\ldots,K\}$, the policy samples an action chunk $\mathbf c_{h,k}\sim\pi_{\vartheta_{\mathrm{old}}}(\cdot\mid\widehat{\mathbf x}_{h,k},\mathbf y)$, with $\widehat{\mathbf x}_{0,k}=\mathbf x_0$. The chunk is converted into frame-aligned configurations $\mathbf{A}$ as described above; unlike $\mathbf a_t$ in Section~\ref{sec:method_formulation}, $\mathbf c_{h,k}$ denotes an entire policy action chunk rather than one configuration. \xsimfull{} predicts the corresponding observation segment, whose final frame becomes $\widehat{\mathbf x}_{h+1,k}$. The imagined trajectories $\widehat{\tau}_k=(\mathbf x_0,\mathbf c_{0,k},\widehat{\mathbf x}_{1,k},\ldots,\mathbf c_{H-1,k},\widehat{\mathbf x}_{H,k})$ support closed-loop policy optimization without additional RoboTwin interaction.

\paragraph{VLM-based dense reward.}
The frozen evaluator scores $J$ successive prefixes of each rollout, assigning task-completion progress $p_{\phi,j,k}$ and visual validity $a_{\phi,j,k}$ to prefix $j\in\{1,\ldots,J\}$ of trajectory $k$, together with terminal-success confidence $q_{\phi,k}$ for the full trajectory. We define the trajectory return $R_k$ as
\begin{equation}
R_k = \sum_{j=1}^{J}\gamma^{j-1}a_{\phi,j,k}
\left(p_{\phi,j,k}-p_{\phi,j-1,k}\right)
+ \beta a_{\phi,J,k}q_{\phi,k},
\label{eq:policy_improvement_reward}
\end{equation}
where $p_{\phi,0,k}=0$, $\gamma$ discounts progress increments, and $\beta$ weights the terminal-success bonus. This reward combines changes in task completion with terminal feedback, weighted by visual validity. We exclude rollouts below a terminal-validity threshold and resample groups with insufficient valid samples or negligible return variance.

\paragraph{GRPO policy optimization.}
Group Relative Policy Optimization (GRPO)~\citep{shao2024deepseekmath} normalizes returns within each rollout group to obtain trajectory-level advantages $\widehat{A}_k$:
\begin{equation}
\widehat{A}_k = \frac{R_k-\mu_R}{\sigma_R+\epsilon}, \qquad
\mu_R=\frac{1}{K}\sum_{k=1}^{K}R_k,
\label{eq:policy_improvement_advantage}
\end{equation}
where $\mu_R$ and $\sigma_R$ are the within-group mean and standard deviation, respectively, and $\epsilon>0$ ensures numerical stability. Each action chunk receives the trajectory-level advantage. We maximize the clipped objective
\begin{equation}
\mathcal{J}_{\mathrm{GRPO}}(\vartheta)=
\mathbb{E}\!\left[
\frac{1}{KH}\sum_{k=1}^{K}\sum_{h=0}^{H-1}
\min\!\left(
\rho_{h,k}(\vartheta)\widehat{A}_k,
\operatorname{clip}\!\left(\rho_{h,k}(\vartheta),1-\epsilon_{\mathrm{low}},1+\epsilon_{\mathrm{high}}\right)\widehat{A}_k
\right)
\right],
\label{eq:policy_improvement_grpo}
\end{equation}
where $\rho_{h,k}(\vartheta)=\pi_{\vartheta}(\mathbf c_{h,k}\mid\widehat{\mathbf x}_{h,k},\mathbf y)/\pi_{\vartheta_{\mathrm{old}}}(\mathbf c_{h,k}\mid\widehat{\mathbf x}_{h,k},\mathbf y)$ is the action-chunk probability ratio. The operator $\operatorname{clip}$ restricts this ratio to $[1-\epsilon_{\mathrm{low}},1+\epsilon_{\mathrm{high}}]$, with $\epsilon_{\mathrm{low}}$ and $\epsilon_{\mathrm{high}}$ controlling the lower and upper clipping margins. The expectation is over rollout groups sampled from the behavior policy and the training initial conditions and instructions.
% TODO: Specify action-token probability aggregation into chunk probabilities
% and whether resampling restores K valid trajectories before each update.

\paragraph{Baselines and evaluation.}
All variants share the same supervised initialization. We compare the initial SFT policy, Iter-SFT on VLM-filtered successful rollouts, \xsimfull{}-GRPO with binary terminal reward, and the full reward in Equation~\ref{eq:policy_improvement_reward}. Training variants use matched tasks and optimization budgets, with matched world-model rollout budgets where applicable. RoboTwin-GRPO uses simulator task-checker feedback as an oracle reference. Final policies are evaluated in RoboTwin's randomized configuration on shared unseen seeds, using the task checker rather than the training VLM. We report per-task and mean success rates across five tasks, averaged over three independent policy-training seeds.

Table~\ref{tab:policy_improvement} shows that the policy trained with the full world-model--VLM optimization pipeline achieves a mean success rate of $75.4\%$, exceeding initial SFT, Iter-SFT, and terminal-only GRPO by $12.7$, $8.4$, and $6.9$ percentage points, respectively. Gains over initial SFT occur on all five tasks, with the largest on T4 ($53.7\%$ to $68.0\%$); the full method remains $8.4$ points below RoboTwin-GRPO. The terminal-only comparison evaluates the combined reward design, rather than isolating individual progress and validity contributions.

% Required package: booktabs
% TODO: Replace T1--T5 with the corresponding RoboTwin task names.
% Reward-component ablations can be moved to a separate appendix table if needed.

\begin{table}[tbp]
    \centering
    \normalsize
    \caption{Policy improvement on five RoboTwin tasks under the randomized configuration. Success rates are measured on shared unseen seeds and averaged over three policy-training runs; RoboTwin-GRPO is a simulator-based oracle reference.}
    \label{tab:policy_improvement}
    \normalsize
    \setlength{\tabcolsep}{4pt}
    \begin{tabular*}{\linewidth}{@{\extracolsep{\fill}}p{0.40\linewidth}cccccc@{}}
        \toprule
        \multirow{2}{*}{\textbf{Method}}
        & \multicolumn{6}{c}{\textbf{RoboTwin success (\%) $\uparrow$}} \\
        \cmidrule(lr){2-7}
        & \textbf{T1}
        & \textbf{T2}
        & \textbf{T3}
        & \textbf{T4}
        & \textbf{T5}
        & \textbf{Mean $\uparrow$} \\
        \midrule
        Initial OpenVLA-OFT (SFT)
        & 67.8 & 58.3 & 71.2 & 53.7 & 62.5 & 62.7 \\
        Iter-SFT (VLM-filtered)
        & 71.5 & 62.7 & 74.8 & 58.8 & 67.3 & 67.0 \\
        RoboTwin-GRPO (oracle)
        & 88.2 & 79.5 & 90.0 & 76.8 & 84.3 & 83.8 \\
        \midrule
        \xsimfull{}-GRPO (terminal only)
        & 72.8 & 63.4 & 76.2 & 59.7 & 70.4 & 68.5 \\
        \textbf{\xsimfull{}-GRPO (full)}
        & \textbf{79.2} & \textbf{71.0} & \textbf{82.3} & \textbf{68.0} & \textbf{76.5} & \textbf{75.4} \\
        \bottomrule
    \end{tabular*}
\end{table}

% TODO: Report K, H, J, gamma, beta, validity threshold, GRPO clipping
% coefficients, sampling temperature, learning rate, and optimization steps.
% TODO: Replace T1--T5 with task names and report evaluation episodes per task,
% variability across training seeds, and the binary terminal-reward threshold.

\subsection{Generalization under Distribution Shifts}
\label{sec:ood_generalization}

A general-purpose world-model simulator should accurately capture action-conditioned dynamics beyond its training distribution. We therefore evaluate \xsimfull{} along five controlled axes: action trajectory, scene appearance, robot embodiment, manipulated object, and camera viewpoint. The task semantics and action horizon remain fixed while the designated factor is varied.

\begin{figure}[tbp]
    \centering
    \begingroup
    \ifdefined\reportincludegraphics\let\includegraphics\reportincludegraphics\fi
    \includegraphics[width=\linewidth]{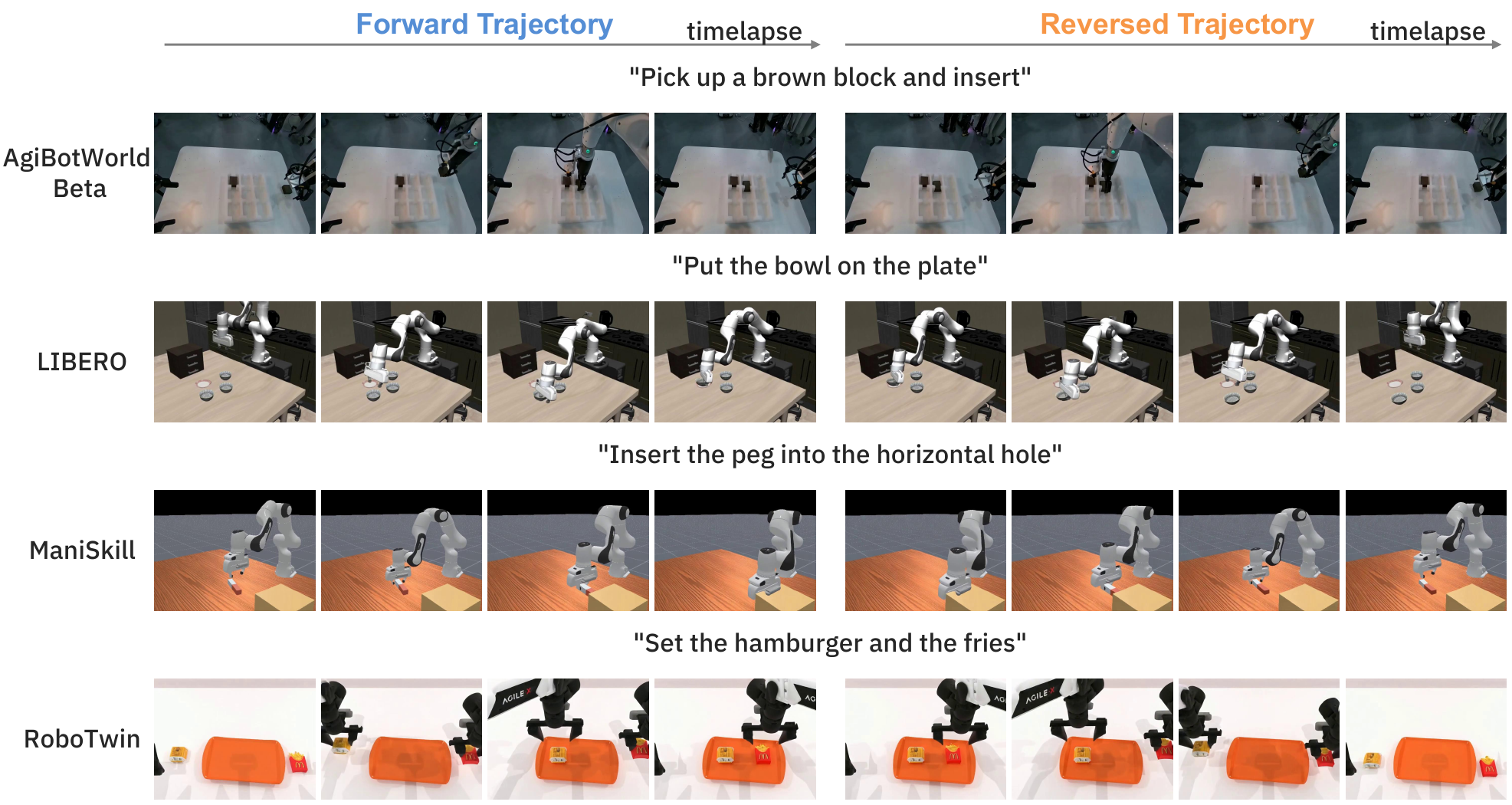}
    \endgroup
    \caption{
        \textbf{Trajectory OOD.}
        Forward and reversed action-conditioned rollouts across four real-world and simulated benchmarks.
    }
    \label{fig:trajectory_ood}
\end{figure}

\paragraph{Trajectory OOD.}
We evaluate trajectory-level generalization using reversed robot motions. Given a forward rollout, we reverse its frame-aligned robot-configuration sequence and use the terminal observation as the new conditioning frame. As shown in Figure~\ref{fig:trajectory_ood}, the generated motions remain consistent with the reversed commands across both real-world and simulated tasks, driving the robot toward its original initial configuration. This setting assesses the sensitivity of the model to the supplied trajectory rather than to dominant forward-motion patterns in the training data. It evaluates action-conditioned trajectory reversal without assuming exact time reversibility in contact-rich dynamics.

\begin{figure}[tbp]
    \centering
    \begingroup
    \ifdefined\reportincludegraphics\let\includegraphics\reportincludegraphics\fi
    \includegraphics[width=\linewidth]{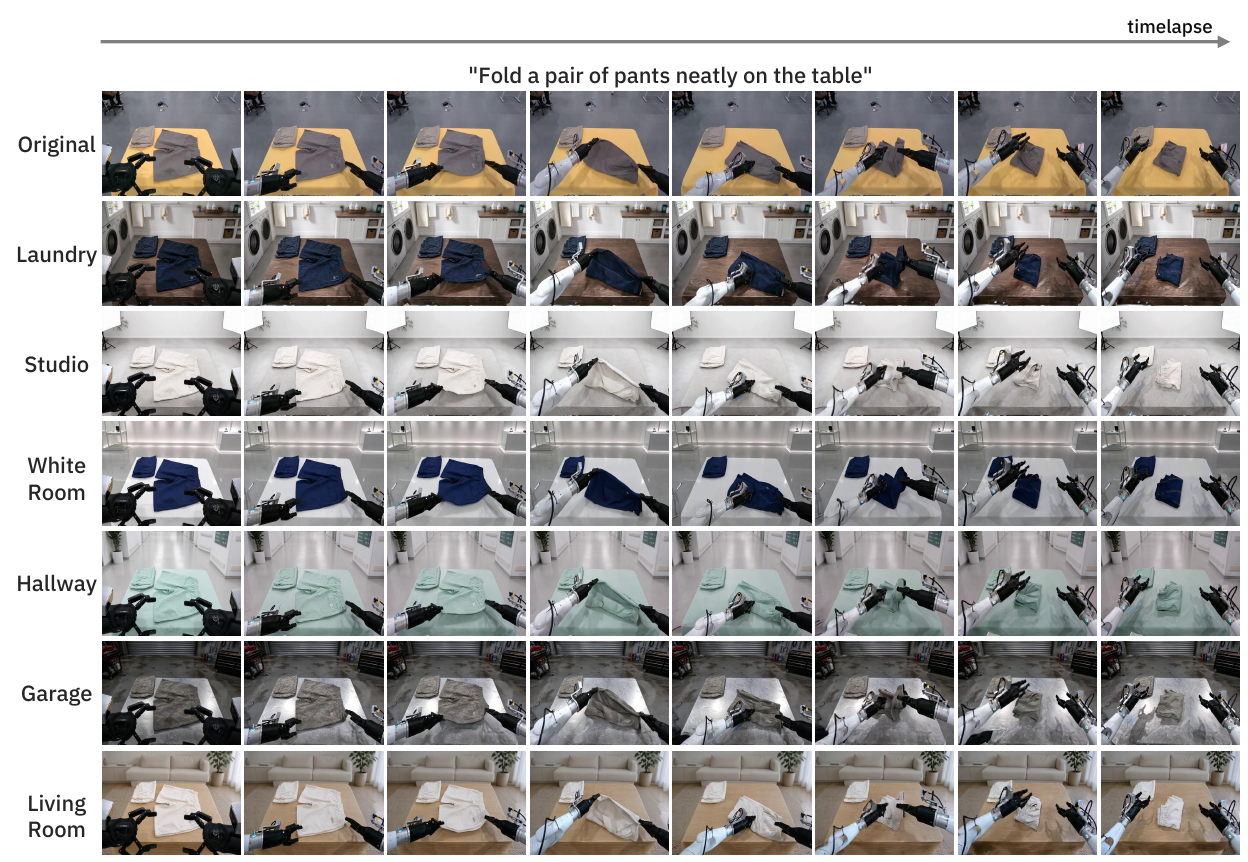}
    \endgroup
    \caption{
        \textbf{Scene/appearance OOD.}
        Pants-folding rollouts in the original scene and six image-edited environments.
    }
    \label{fig:scene_appearance_ood}
\end{figure}

\paragraph{Scene/appearance OOD.}
We use GPT-Image-2~\citep{openai2026gptimage2} to synthesize OOD scene variants from the conditioning frame while preserving the task, robot pose, and spatial layout, and then reuse the same action trajectory for generation. Figure~\ref{fig:scene_appearance_ood} includes six edited variants in addition to the original scene. Despite substantial changes in background, illumination, texture, and local object appearance, the generated rollouts remain temporally coherent and consistent with the specified folding behavior. This experiment evaluates broad scene and appearance variation rather than an isolated background shift.

\begin{figure}[tbp]
    \centering
    \begingroup
    \ifdefined\reportincludegraphics\let\includegraphics\reportincludegraphics\fi
    \includegraphics[width=\linewidth]{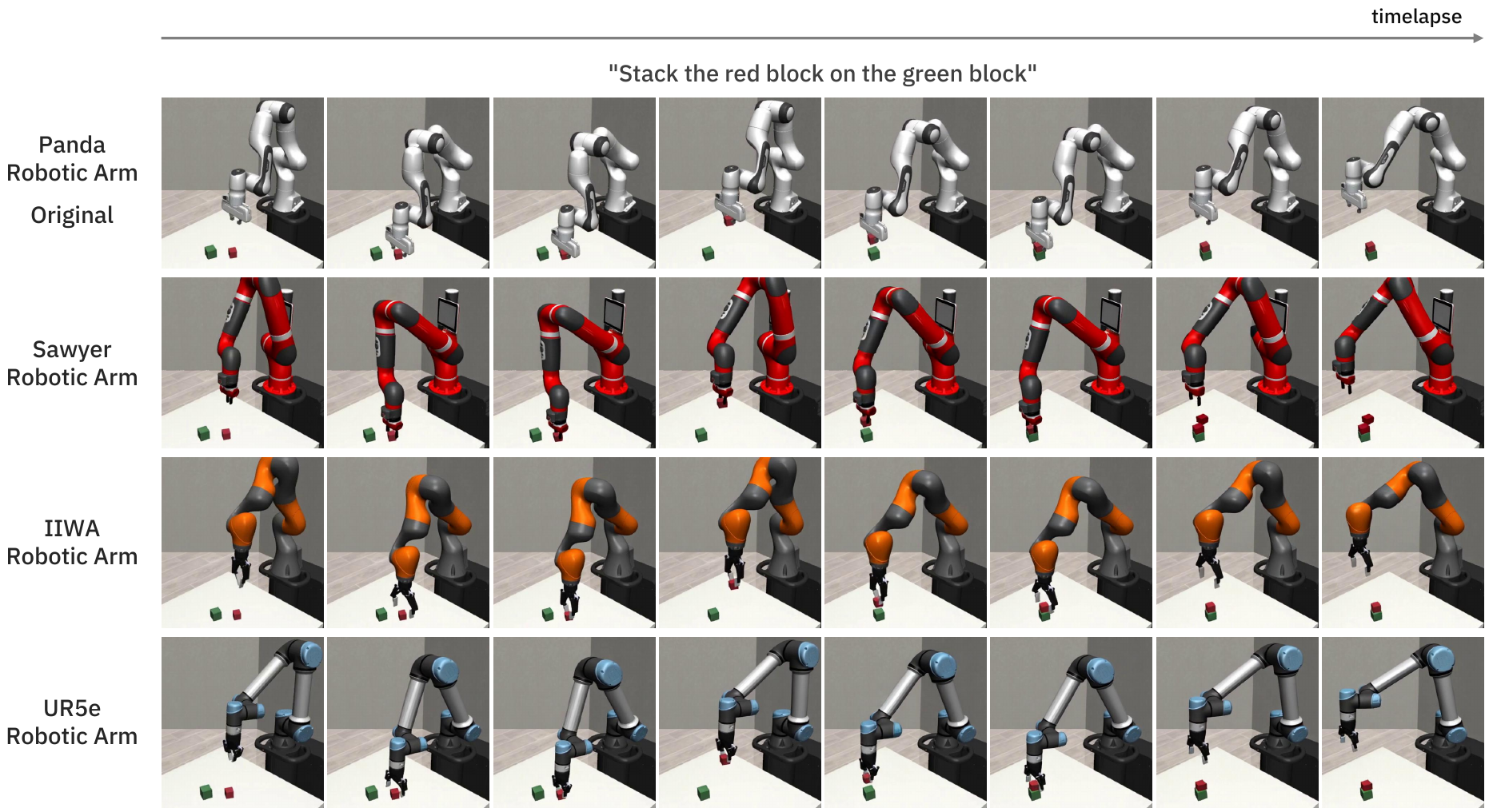}
    \endgroup
    \caption{
        \textbf{Robot-embodiment OOD.}
        Block-stacking rollouts retargeted from Panda to held-out Sawyer, IIWA, and UR5e embodiments.
    }
    \label{fig:embodiment_ood}
\end{figure}

\paragraph{Robot-embodiment OOD.}
To assess embodiment-level generalization, we retarget a manipulation trajectory from a source robot to embodiments excluded from training. Specifically, we extract the Cartesian end-effector pose trajectory of the source execution and solve inverse kinematics for each target robot, yielding embodiment-specific joint trajectories that are subsequently converted into unified action values and camera-aligned action videos. As shown in Figure~\ref{fig:embodiment_ood}, \xsimfull{} generates block-stacking motions consistent with the source Panda trajectory for unseen Sawyer, IIWA, and UR5e embodiments. The qualitative results suggest that the model retains task-level motion semantics while accommodating changes in robot kinematics and projected geometry.

\begin{figure}[tbp]
    \centering
    \begingroup
    \ifdefined\reportincludegraphics\let\includegraphics\reportincludegraphics\fi
    \includegraphics[width=\linewidth]{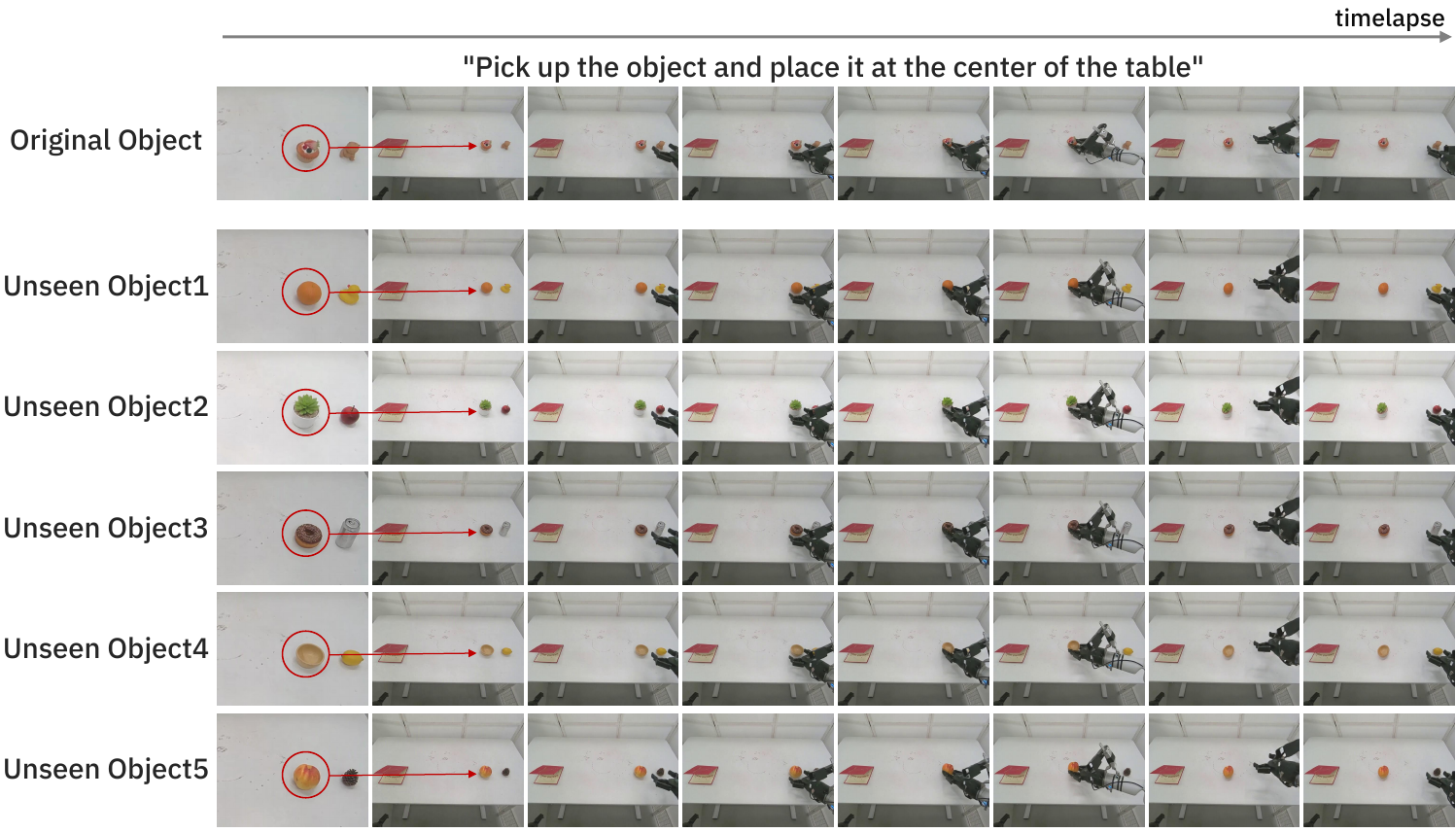}
    \endgroup
    \caption{
        \textbf{Object OOD.}
        Pick-and-place rollouts for the original object and five image-edited object variants.
    }
    \label{fig:object_ood}
\end{figure}

\paragraph{Object OOD.}
We replace the manipulated object in the initial observation using GPT-Image-2~\citep{openai2026gptimage2}, while preserving the robot configuration, coarse task-relevant spatial relations, and action trajectory. Figure~\ref{fig:object_ood} compares the original object with five edited variants that differ in appearance, shape, and scale. Given the same pick-and-place command, the generated motions remain directed toward the edited object, move it toward the center of the table, and largely preserve its visual identity over time. These observations suggest reduced reliance on the appearance of specific training objects. This setting represents controlled visual object variation and does not assume that the edited objects share identical physical properties.

\begin{figure}[tbp]
    \centering
    \begingroup
    \ifdefined\reportincludegraphics\let\includegraphics\reportincludegraphics\fi
    \includegraphics[width=\linewidth]{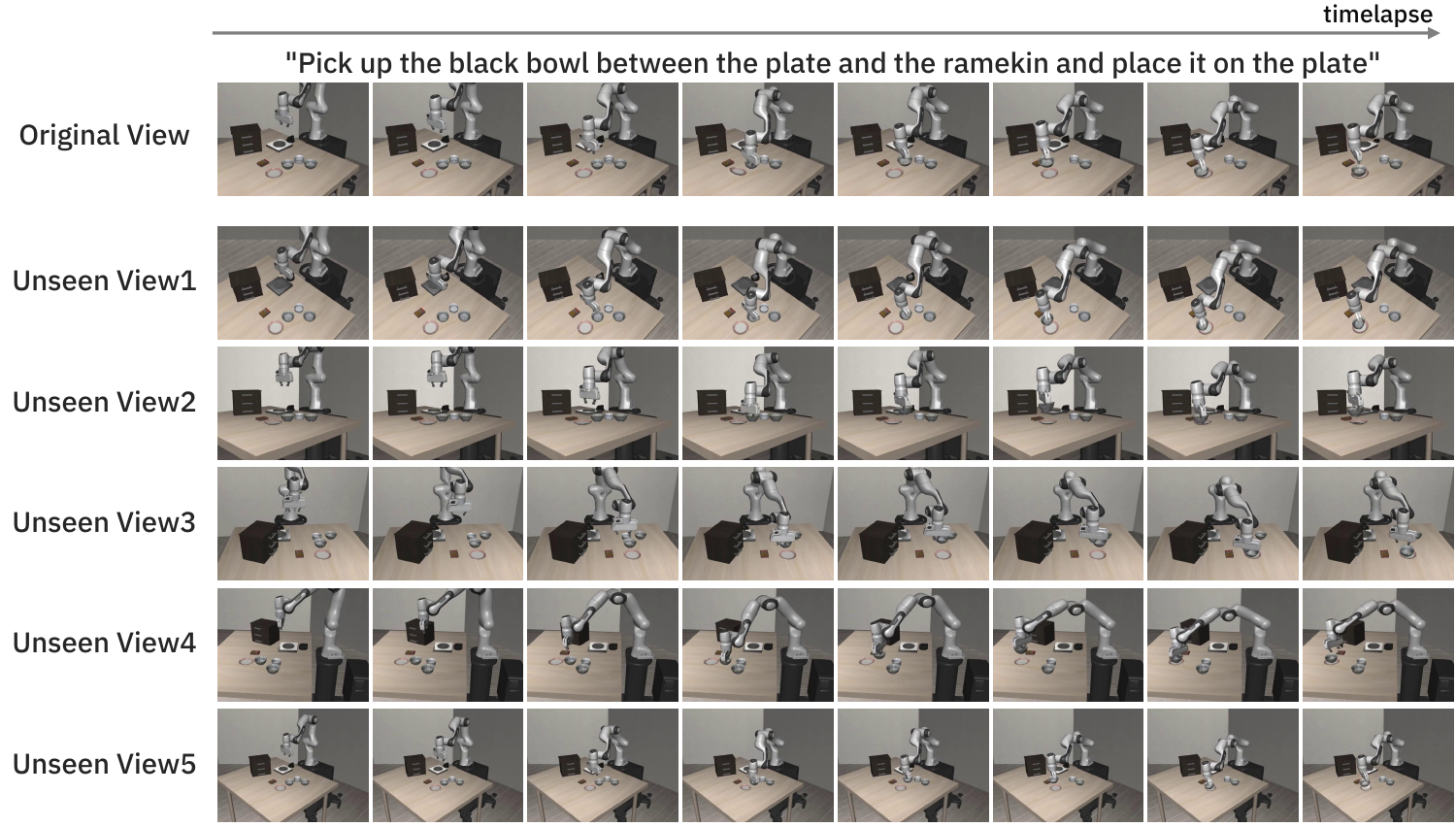}
    \endgroup
    \caption{
        \textbf{Viewpoint OOD.}
        Action-conditioned rollouts from the original camera and five unseen viewpoints.
    }
    \label{fig:viewpoint_ood}
\end{figure}

\paragraph{Viewpoint OOD.}
We further evaluate initial observations captured from camera viewpoints excluded from training. For each target view, the corresponding camera calibration is used to render an action video spatially aligned with the novel viewpoint. Figure~\ref{fig:viewpoint_ood} compares the original camera with five unseen viewpoints for the task of picking up the black bowl between the plate and the ramekin and placing it on the plate. Across substantial changes in elevation, azimuth, and distance, the generated sequences remain consistent with the commanded robot motion and the intended interaction, suggesting robustness to camera configurations not observed during training.

Taken together, these controlled studies demonstrate broad qualitative generalization across temporal, visual, embodiment, and geometric distribution shifts. The results suggest that the unified numerical and visual action representations capture transferable motion structure, supporting \xsimfull{} as a general-purpose world-model simulator for diverse robotic manipulation settings.

% Remaining experiment floats may share pages with the conclusion; the
% entry points flush them before the bibliography.

\section{Conclusion and Future Work}
\label{sec:conclusion}

We presented \xsimfull{}, an action-conditioned world-model simulator for robotic manipulation across heterogeneous embodiments and environments. \xsimfull{} combines a unified 28-dimensional robot-configuration representation with URDF-rendered action videos, providing precise numerical specifications and explicit image-space motion guidance through complementary conditioning pathways in a Video DiT with sparse MoE layers for modeling heterogeneous robot dynamics. A multi-stage causal distillation procedure further reduces sampling from 35 to four steps for rollout-intensive applications. Experiments on held-out test splits from real-world and simulated datasets demonstrate improved video quality and action consistency and validate the contributions of the action representations, mixed-domain training, and the MoE design. Generated rollouts support data generation for policy training, while their integration with a fine-tuned VLM evaluator supports policy evaluation and ranking, action selection, and policy improvement. In these downstream procedures, the simulator predicts visual outcomes, the evaluator supplies task-conditioned feedback, and selection or optimization algorithms use that feedback to choose actions or update policies. The model also retains task-consistent motion under controlled changes in trajectory, scene appearance, embodiment, object identity, and viewpoint, positioning \xsimfull{} as a practical step toward a general-purpose world-model simulator for embodied intelligence.

Future work will pursue three directions. First, we aim to expand the curated training corpus to cover a wider range of embodiments, object interactions, sensing conditions, and long-horizon behaviors, and assess how this broader coverage affects transfer across domains. Second, we will systematically characterize the quality-efficiency trade-offs of causal adaptation and few-step distillation, particularly their effects on fine-grained dynamics, rollout stability, and generalization. Adaptive computation and hybrid teacher-student inference are potential approaches to reducing latency while preserving predictive quality. Third, we aim to make \emph{generality} more precisely measurable through evaluation protocols that distinguish within-source prediction from transfer under explicitly defined distribution shifts. Such protocols should assess generalization across tasks, embodiments, action spaces, environments, and temporal horizons using complementary measures of perceptual fidelity, action consistency, physical plausibility, and downstream utility. Together, these directions would help establish the capabilities and limitations of general-purpose world-model simulators beyond broad training-data coverage.
\clearpage
% Keep result figures out of the bibliography without forcing early page ends
% at the qualitative or OOD subsection boundaries.
\FloatBarrier
\bibliographystyle{iclr2027_conference}
\bibliography{iclr2027_conference}

% Keep the short contributor block together, sharing the bibliography's last
% page when space permits rather than always forcing a new page.
\par\noindent
\begin{minipage}{\linewidth}
    % Core-contributor responsibilities supplied by the authors.
% Other contributor names and roles are retained from the report template.
\vspace{10pt} 
\section{Contributions}
\label{sec:contributions}
Contributors and their roles in the development of \xsimfull{} are listed below.
\vspace{10pt} 
\subsection{Core Contributors}
\noindent\textbf{Shilong Zou} contributed to data preparation, model architecture design, world-model training, and policy training for downstream tasks.

\noindent\textbf{Shilin Zhang} contributed to data preparation, policy training for downstream tasks, and visualization of experimental results.
\vspace{10pt} 
\subsection{Contributors}
Yingji Zhang, Yuhang Huang, Yi Zhang, Zeyuan Ding, Han Dong, Junwei Liao
\vspace{10pt} 
\subsection{Tech Lead}
Yong Dai, Shilong Zou
\vspace{10pt} 
\subsection{Corresponding Authors}
Jian Tang, Xiaozhu Ju
\end{minipage}
\end{document}